\documentclass[10pt,twocolumn,letterpaper]{article}

\usepackage[pagenumbers]{cvpr} 

\usepackage{newfloat}
\usepackage{listings}
\usepackage{amssymb} %
\usepackage{booktabs}
\usepackage{multirow} %
\usepackage{amsmath}
\usepackage{pifont}
\usepackage{xcolor} 
\usepackage[utf8]{inputenc} %
\usepackage[T1]{fontenc}    %
\usepackage{url}            %
\usepackage{amsfonts}       %
\usepackage{nicefrac}       %
\usepackage{microtype}      %
\usepackage{graphicx}
\usepackage[normalem]{ulem}
\usepackage{rotating}
\usepackage{makecell}
\usepackage{stfloats}
\usepackage{wrapfig}
\usepackage{enumitem}
\usepackage{scalerel}
\usepackage{color}
\usepackage[dvipsnames]{xcolor}
\usepackage{scalerel}
\usepackage{colortbl}

\usepackage{algorithm} %
\usepackage{algorithmic} %

\definecolor{third}{rgb}{1, 1, 0.7}
\definecolor{second}{rgb}{1, 0.85, 0.7}
\definecolor{first}{rgb}{1, 0.7, 0.7}
\newcommand{\fs}{\cellcolor{first}}   
\newcommand{\nd}{\cellcolor{second}}      
\newcommand{\rd}{\cellcolor{third}}      

\definecolor{cvprblue}{rgb}{0.21,0.49,0.74}
\definecolor{shapecolor}{rgb}{0.0,0.5,0.0}
\usepackage[pagebackref,
breaklinks,
colorlinks, 
citecolor=cvprblue,
linkcolor=red,
anchorcolor=red,
]{hyperref}

\def\paperID{1313} 
\def\confName{CVPR}
\def\confYear{2025}

\newcommand{\name}{JamMa}
\title{JamMa: Ultra-lightweight Local Feature Matching with Joint Mamba}

\author{Xiaoyong Lu \qquad Songlin Du\footnotemark[1]\\
School of Automation, Southeast University, Nanjing, China \\
{\tt\small \{luxiaoyong, sdu\}@seu.edu.cn}
}

\begin{document}

\twocolumn[{%
    \renewcommand\twocolumn[1][]{#1}%
    \maketitle
	\begin{center}
	\vspace{-7mm}
	\includegraphics[width=0.98\linewidth]{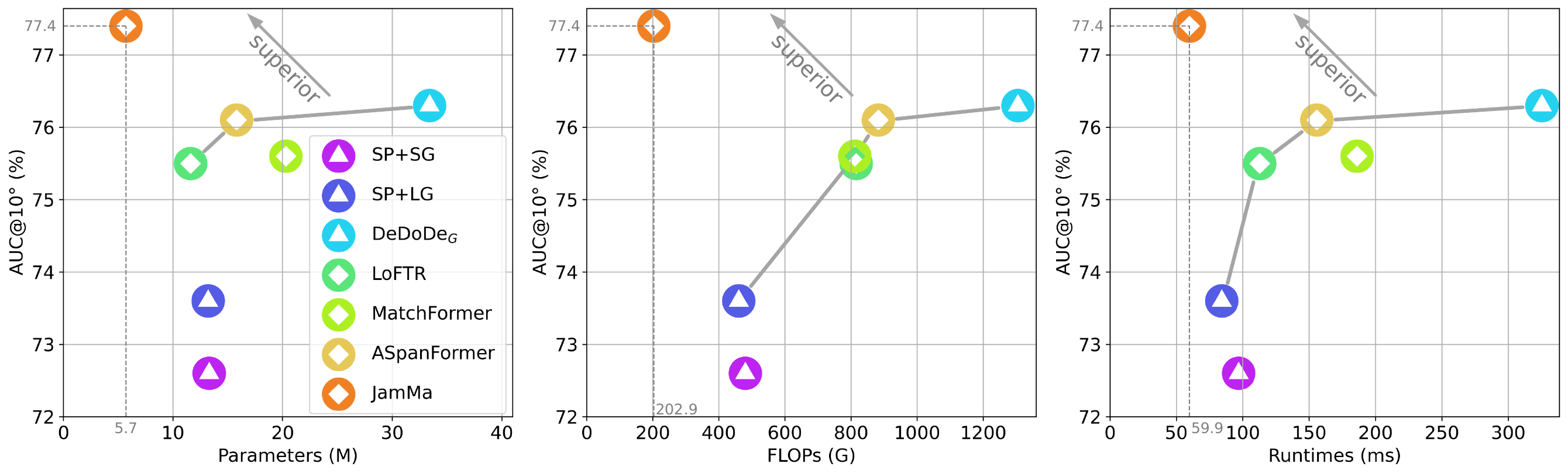}
	\vspace{-3mm}
	\captionof{figure}{
		\textbf{Efficiency vs. Performance.} State-of-the-art sparse \raisebox{-2pt}{\includegraphics[width=10pt]{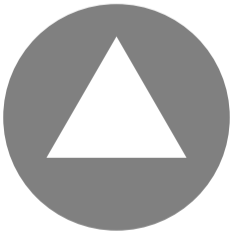}} and 
		semi-dense \raisebox{-2pt}{\includegraphics[width=10pt]{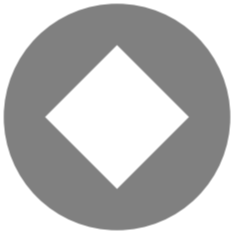}} methods are compared in the MegaDepth dataset \cite{megadepth}. 
		In all three commonly used efficiency metrics, the proposed \name{} achieves a superior performance-efficiency balance by a clear margin.
		}
	\label{figure1}
	\vspace{3mm}
	\end{center}
}]

\footnotetext[1]{\label{corrauth}Corresponding author.}

\begin{abstract}
Existing state-of-the-art feature matchers capture long-range dependencies with Transformers but are hindered by high spatial complexity,
leading to demanding training and high-latency inference.
Striking a better balance between performance and efficiency remains a challenge in feature matching.
Inspired by the linear complexity $\mathcal{O}(N)$ of Mamba, 
we propose an ultra-lightweight Mamba-based matcher, named JamMa, 
which converges on a single GPU and achieves an impressive performance-efficiency balance in inference.
To unlock the potential of Mamba for feature matching,
we propose Joint Mamba with a scan-merge strategy named $\textbf{JEGO}$, which enables:
(1) $\textbf{J}$oint scan of two images to achieve high-frequency mutual interaction, (2) $\textbf{E}$fficient scan with skip steps to reduce sequence length, 
(3) $\textbf{G}$lobal receptive field, and (4) $\textbf{O}$mnidirectional feature representation.
With the above properties, the JEGO strategy significantly outperforms the scan-merge strategies proposed in VMamba and EVMamba in the feature matching task.
Compared to attention-based sparse and semi-dense matchers, 
JamMa demonstrates a superior balance between performance and efficiency,
delivering better performance with less than $50\%$ of the parameters and FLOPs.
Project page: \url{https://leoluxxx.github.io/JamMa-page/}.
\end{abstract}    
\section{Introduction}
\label{sec:intro}
Feature matching, which seeks to establish accurate correspondences between two images, 
underpins various critical tasks, including structure from motion (SfM) \cite{sfm}
and simultaneous localization and mapping (SLAM) \cite{slam}. 
However, feature matching faces challenges arising from variations in viewpoint, illumination, and scale.
Furthermore, applications that demand real-time performance place significant pressure on the efficiency of the matching process.

Feature matching methods are generally classified into three categories: sparse, semi-dense, and dense methods.
Sparse methods \cite{superglue, sgmnet, clustergnn, paraformer}, 
which operate alongside a keypoint detector \cite{sift, superpoint}, 
are designed to establish correspondences between two sets of keypoints.
Therefore, when the detector performs poorly in challenging scenes, such as texture-less scenes, it will severely affect the sparse matcher.
In contrast, semi-dense and dense methods \cite{loftr, aspanformer, dkm, roma} directly establish correspondences between grid points or all pixels, 
enhancing robustness in challenging scenes.
However, semi-dense and dense matchers typically rely on Transformers \cite{transformer,linear_attn} to model long-range dependencies in dense features, 
resulting in impractically high complexity when processing high-resolution images.
Achieving an optimal balance between performance and efficiency remains a critical challenge in feature matching.

To achieve high efficiency and robustness in challenging scenes, 
we explore the potential of accelerating semi-dense paradigm with Mamba,
which captures long-range dependencies with linear complexity and has demonstrated significant success in natural language processing (NLP).
However, since Mamba is originally designed as a sequence model for NLP, 
we identify three key challenges that hinder its suitability for feature matching and propose the Joint Mamba with JEGO strategy to address these issues efficiently:
\begin{table}[t]
	\centering
	\resizebox{0.95\linewidth}{!}{
	\begin{tabular}{l | c c c}
	\toprule
	Method          &Omnidirectional  &Global      &Complexity \\
	\midrule
	Transformer \cite{transformer}     &-           &\ding{51}      &~~$N^{2}$ \\
	EVMamba \cite{efficientvmamba} &\ding{55}   &\ding{55}      &$N$ \\
    Vim \cite{vim}             &\ding{55}   &\ding{51}      &$2N$ \\
	VMamba \cite{vmamba}          &\ding{51}   &\ding{51}      &$4N$\\
    \name{}           &\ding{51}   &\ding{51}      &$N$ \\
	\bottomrule
	\end{tabular}
	}
	\vspace{-2mm}
	\caption{\textbf{Model Property.} $N$ denotes the number of features.}
	\label{tab:Model Properties Analysis}
	\vspace{-6mm}
\end{table}

\textbf{(1) Lack of Mutual Interaction.}
Mamba is designed for modeling a single set of features, \ie, \emph{internal interaction},
whereas feature matching inherently requires interaction between two sets of features, \ie, \emph{mutual interaction}. 
Therefore, a new scan strategy is needed to effectively establish cross-view dependencies between two images.
To this end, we explore two candidate strategies: joint scan and sequential scan, as shown in \cref{joint}.
Our findings indicate that joint scan, which alternates between the two images to facilitate \emph{high-frequency mutual interaction}, 
significantly outperforms the sequential scan, in which each image is scanned individually and in sequence.
Considering the necessity of mutual interaction for feature matching, 
the joint scan is adopted as a core component of the JEGO strategy, 
which also gives rise to the term Joint Mamba.

\textbf{(2) Unidirectionality.}
Mamba is unidirectional as it generates sequences with a 1D forward scan.
However, features exhibit relationships in four basic directions in 2D images, requiring the development of new scan strategies.
As shown in \cref{properties}, Vim \cite{vim} and VMamba \cite{vmamba} propose bi-directional and four-directional scans, respectively, 
but at the cost of a $2\times$ and $4\times$ increase in total sequence length.
EVMamba \cite{efficientvmamba} maintains a total sequence length equal to the original feature number $N$
by skipping during scanning
but fails to achieve omnidirectionality.
The proposed JEGO strategy reconciles the skip scan with the four-directional scan,
achieving both omnidirectionality and a total sequence length of $N$.

\textbf{(3) Causality.} 
Mamba operates as a causal model, allowing each feature in the sequence to perceive only the preceding features.
As a result, only the end \raisebox{-2pt}{\includegraphics[width=10pt]{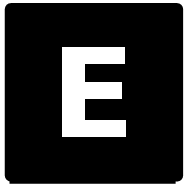}} of the sequence has access to the global receptive field.
Numerous works have validated the importance of large receptive fields for visual models, including visual Mamba models.
As shown in \cref{properties}, Vim \cite{vim} and VMamba \cite{vmamba} achieve a global receptive field through backward scans (dashed lines),
increasing the total sequence length to $2N$ and $4N$.
EVMamba \cite{efficientvmamba} maintains high efficiency but only forward-scans the sequence, 
which limits the global receptive field to only the bottom-right corner.
In the JEGO strategy, the scans in four directions are arranged to ensure a \emph{balanced receptive field}, 
where \emph{small receptive field features are consistently adjacent to those with larger ones} as shown in \cref{properties}.
A CNN termed aggregator is then employed to aggregate global information from four directions into features with small receptive fields. 
We find that a simple aggregator on a balanced receptive field has a surprisingly effective impact, 
intuitively ensuring that each feature is \emph{global} and \emph{omnidirectional}, see \cref{erf}.
To summarize, this paper presents the following contributions:
\begin{figure}[t]
	\includegraphics[width=0.9\linewidth]{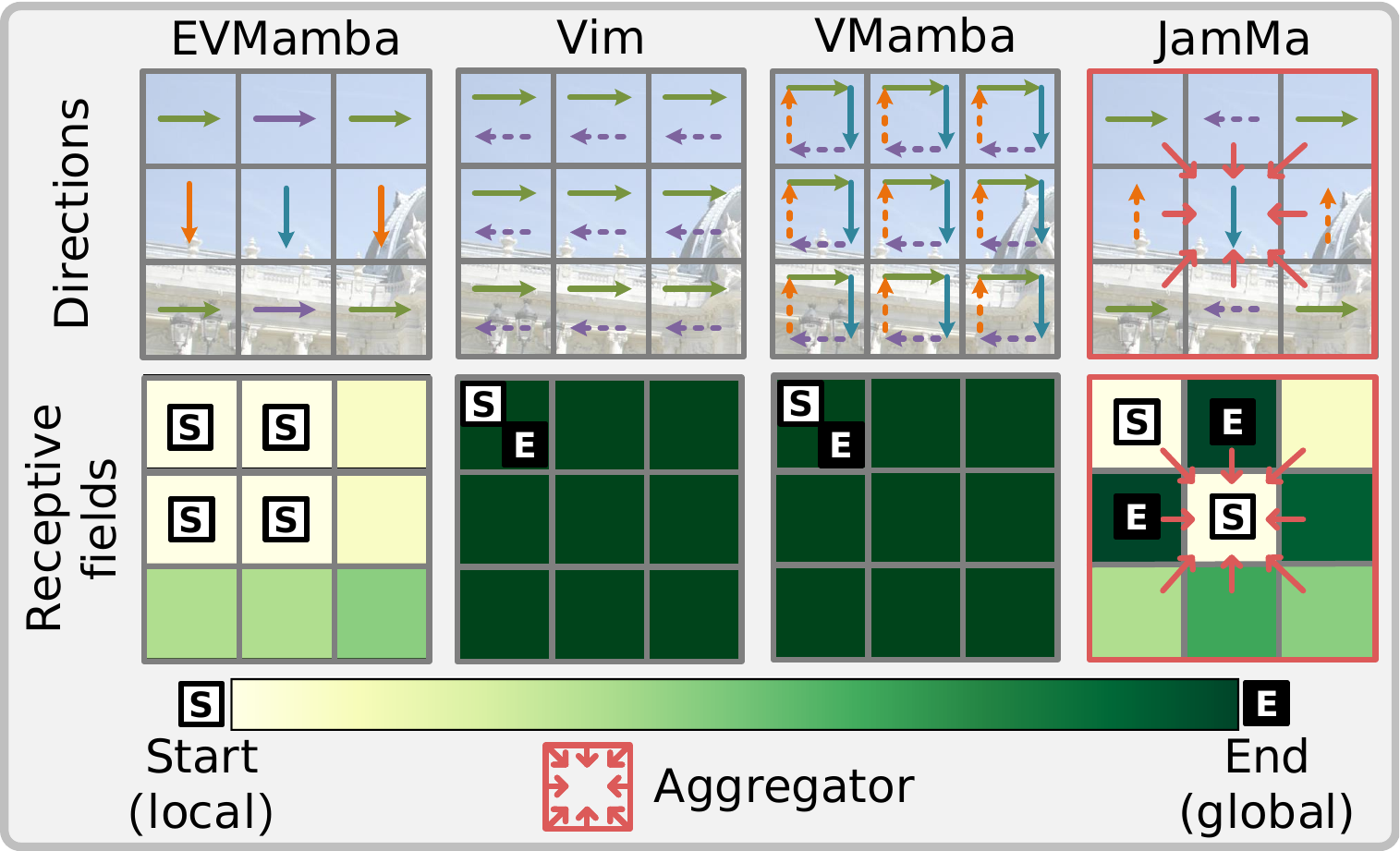}
	\centering
    \vspace{-2mm}
	 \caption{
        \textbf{Receptive Fields and Sequence Directions of Visual Mamba Models.}
	 }
	 \label{properties}
     \vspace{-6mm}
\end{figure}

\begin{itemize}
    \item We present \name{}, an ultra-lightweight semi-dense feature matcher based on Joint Mamba with JEGO strategy.
    \item A scan-merge strategy named \textbf{JEGO} is proposed, 
    which \textbf{J}ointly scans cross-view features with high \textbf{E}fficiency and produces \textbf{G}lobal \textbf{O}mnidirectional features.
    \item High-frequency mutual interaction and a local aggregator on a balanced receptive field yield surprisingly robust features with minimal computational overhead.
    \item Quantitative and qualitative experiments demonstrate the excellent performance-efficiency balance of \name{} and highlight the critical role of the JEGO strategy.
\end{itemize}
\section{Related Works}
\label{sec:related}
\subsection{Local Feature Matching}

\noindent\textbf{Sparse matching} encompasses well-known handcrafted methods such as SIFT \cite{sift}, SURF \cite{surf}, BRIEF \cite{brief}, and ORB \cite{orb}.
The rise of deep learning has shifted research focus toward learning-based methods.
Models like R2D2 \cite{r2d2} and SuperPoint \cite{superpoint} employ convolutional neural networks (CNN) to improve robustness.
SuperGlue \cite{superglue} is the first to model global relationships between sparse features using Transformers \cite{transformer}, 
enriching features through self- and cross-attention.
Recent attention-based sparse methods \cite{sgmnet, clustergnn, paraformer,lightglue} generally address the computational inefficiencies
but still rely heavily on the performance of detector.

\noindent\textbf{Semi-dense matching} methods are pioneered by LoFTR \cite{loftr}, 
which proposes a coarse-to-fine paradigm to establish coarse matches between grids and then adjust the matching points with fine features.
ASpanFormer \cite{aspanformer} proposes a new alternative to linear attention \cite{linear_attn} that adaptively adjusts the span of attention based on uncertainty.
MatchFormer \cite{matchformer} abandons the CNN-based feature extractor and proposes a purely attention-based matcher to improve robustness.
Semi-dense methods excel in texture-less scenes because they match grids distributed across the image.
However, the large number of grid points also introduces a greater computational burden, 
particularly when matchers utilize attention mechanisms to establish long-distance dependencies.

\noindent\textbf{Dense matching} methods directly estimate dense warp to establish correspondence for each pixel between two images.
Some approaches \cite{ncnet, dual, 9320451} utilize the 4D correlation volume to predict the dense warp.
Recently, DKM \cite{dkm} and RoMa \cite{roma} have introduced a kernelized global matcher and an embedding decoder to initially predict coarse warps, which are subsequently refined using stacked feature maps.
While DKM and RoMa outperform sparse and semi-dense methods in terms of accuracy and robustness, they demand significantly more parameters and longer runtimes.

\begin{figure*}[t]
	\includegraphics[width=0.99\linewidth]{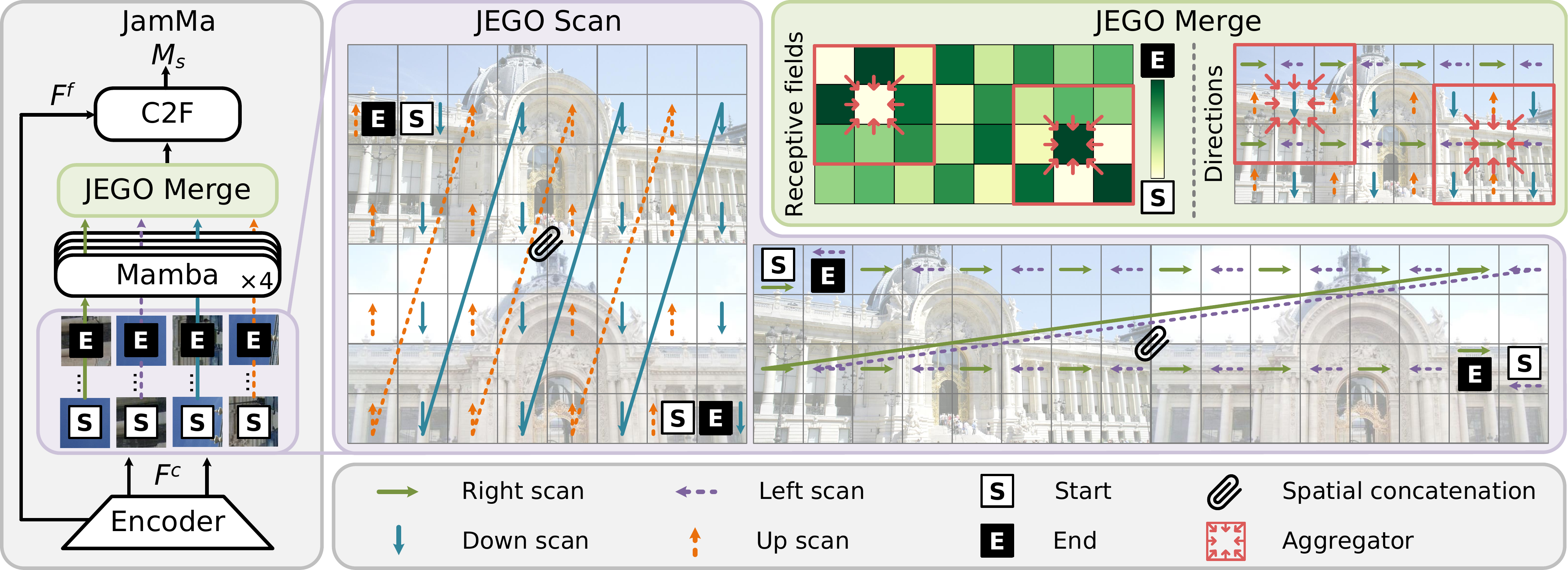}
	\centering
	\vspace{-2mm}
	 \caption{
    \textbf{Overview of the Proposed Method.} 
	\name{} extracts coarse and fine local features with a CNN encoder (\cref{subsec:encoder}) and scans the coarse features with the JEGO scan module (\cref{subsec:scan}).
	The four sequences are processed by four independent Mamba blocks and then merged back into 2D feature maps by the JEGO merge module (\cref{subsec:merge}).
	Finally, the coarse-to-fine matching module (C2F) generates the matching results (\cref{subsec:c2f}).
	We show how the \textbf{JEGO} strategy enables \textbf{J}oint, \textbf{E}fficient, \textbf{G}lobal, and \textbf{O}mnidirectional scanning and merging.
	}
	 \label{overview}
	 \vspace{-3mm}
\end{figure*}

\subsection{State Space Models}
\label{subsec:State Space Models}
State Space Models (SSM) are initially designed to model continuous linear time-invariant systems~\cite{lti},
where an input signal $x(t)$ is mapped to its output signal $y(t)$ as
\begin{align}
    h'(t) = Ah(t) + Bx(t),~~~y(t) = Ch'(t).
\end{align}
$A \in \mathbb{R}^{N \times N}$, $B \in \mathbb{R}^{N \times 1}$ and $C \in \mathbb{R}^{1 \times N}$ are SSM parameters. 
To adapt SSM in discrete systems, \eg, sequence-to-sequence tasks,
S4 \cite{s4} proposes to transform SSM parameters to their discretized counterparts using the zero-order hold rule.
However, S4 shares parameters across all time steps, which severely limits its representational capacity. 
Mamba~\cite{mamba} proposes S6 which makes SSM parameters dependent on the input sequence.
This data-dependent property significantly enhances Mamba, making its performance comparable to Transformer.
Meanwhile, the primary reason for the high efficiency of Mamba is that the repetitive computations in SSM can be performed once by a global convolution kernel,
which can be pre-computed via the SSM parameters.

Following the success of Mamba, there has been a surge in applying the framework to computer vision tasks.
Visual Mamba models generally scan 2D feature maps into 1D sequences to model them with Mamba
and then merge the 1D sequences back to 2D feature maps.
Vim \cite{vim} and VMamba \cite{vmamba} propose the bi-directional scan and four-directional scan, respectively, 
extending the originally causal Mamba to achieve a global receptive field.
EVMamba \cite{efficientvmamba} proposes the skip scan to reduce the sequence length thereby significantly improving efficiency.
The above visual Mamba models focus on single-image tasks, \eg, image classification.
In contrast, we propose Joint Mamba, which is designed to jointly model \emph{two images} for feature matching.
\vspace{-1mm}

\section{Methodology}
\label{sec:methodology}

\subsection{\name{} Overview}
The overall architecture of \name{} is illustrated in \cref{overview}.
We extract the coarse features $F^{c}$ and fine features $F^{f}$ with a CNN encoder.
The coarse features are then processed by Joint Mamba, \ie, JEGO scan $\rightarrow$ Mamba $\rightarrow$ JEGO merge, to perform internal and mutual interaction,
perceiving global and omnidirectional information of two images.
Finally, we utilize the coarse-to-fine matching (C2F) module to generate matching results.
Specifically, we establish bi-directional coarse matches
and adjust coarse match points in fine matching and sub-pixel refinement modules \cite{xoftr}.

\begin{figure}[t]
	\includegraphics[width=0.99\linewidth]{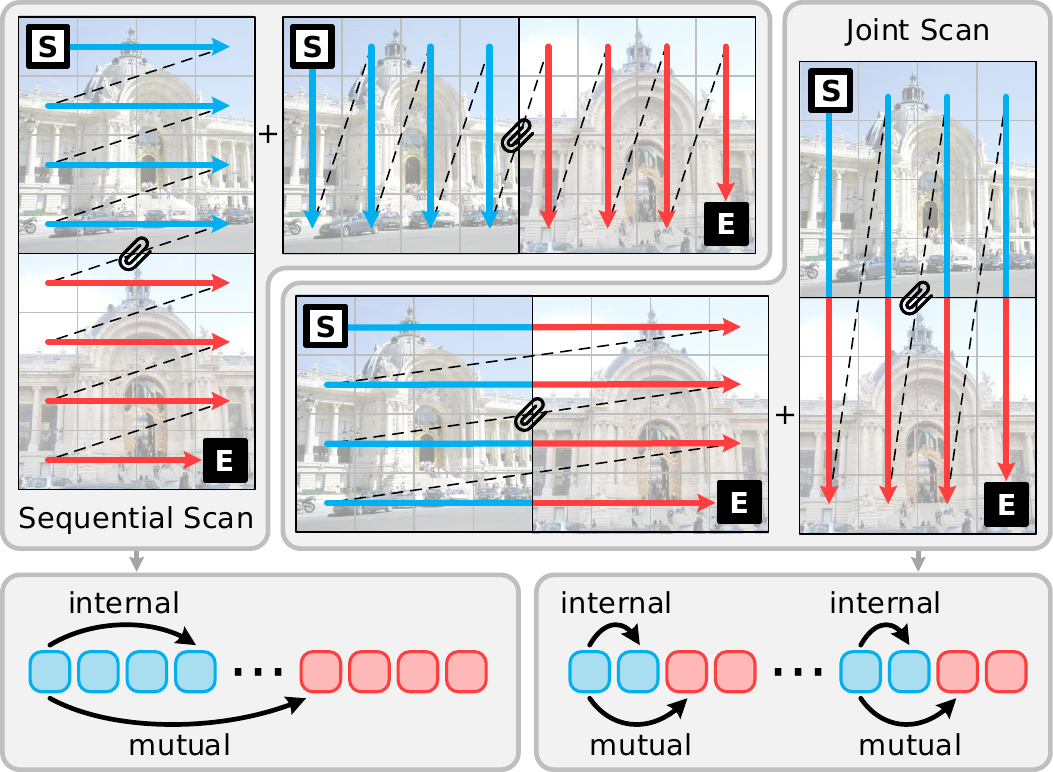}
	\centering
    \vspace{-2mm}
	 \caption{
    \textbf{Sequential scan vs. Joint scan.}
    Sequential scan follows a pattern of internal $\rightarrow$ mutual interaction, 
    similar to self $\rightarrow$ cross attention.
    Joint scan, by contrast, emphasizes high-frequency mutual interaction, 
    which has proven crucial for feature matching.
	 }
	 \label{joint}
     \vspace{-5mm}
\end{figure}

\subsection{Local Feature Extraction}
\label{subsec:encoder}
\vspace{-1mm}
We adopt the ConvNeXt V2 \cite{convnextv2} as the encoder to extract local coarse features $F^{c}\in\mathbb{R}^{H_{c}\times W_{c}\times C_{1}}$ 
and fine features $F^{f}\in\mathbb{R}^{H_{f}\times W_{f}\times C_{2}}$ from images $I_{A}$ and $I_{B}$ as
\begin{align}
    \label{encoder}
    &F^{c}_{A}, F^{f}_{A} = \mathrm{Encoder}(I_{A}), 
    &F^{c}_{B}, F^{f}_{B} = \mathrm{Encoder}(I_{B}).
\end{align}
We observe that the designs of ConvNeXt perform remarkably well in lightweight settings.
Specifically, a ConvNeXt-based encoder with $0.65$M parameters is sufficient to support \name{} for competitive performance.

\subsection{Joint Mamba}
Joint Mamba, comprising JEGO scan, Mamba, and JEGO merge, is proposed to efficiently establish global dependencies in coarse features $F^{c}_{A,B}$ for robust coarse matching.
\vspace{-2mm}
\subsubsection{JEGO Scan}
\label{subsec:scan}
\noindent\textbf{Joint Scan.}
Existing Mamba models are designed for single-image tasks, 
whereas feature matching requires interaction between two sets of features.
Attention-based matchers, for example, utilize self- and cross-attention for internal and mutual interaction.
Therefore, a new scan strategy is needed to achieve both internal and mutual interaction.

We begin by concatenating the coarse features $F^{c}_{A,B}$ both horizontally and vertically as
\vspace{-3mm}
\begin{equation} \label{eq:concat}
    \begin{aligned}
    &X^{h} = [F^{c}_{A}|F^{c}_{B}], 
    &X^{v} = \begin{bmatrix} F^{c}_{A} \\ F^{c}_{B} \end{bmatrix}. 
    \end{aligned}
\end{equation}
\vspace{-4mm}

Based on $X^{h}$ and $X^{v}$, we explore two candidate scan strategies: sequential scan and joint scan, as shown in \cref{joint}.
In the sequential scan, all features of one image are scanned before moving to the next.
In contrast, the joint scan alternates between the two images to enable high-frequency mutual interaction, which is beneficial for cross-view dependencies.
We show in \cref{tab:exp ablation} that the joint scan is significantly more effective for feature matching than the sequential scan.

\noindent\textbf{Efficient Four-directional Scan.}
In addition to the proposed joint scan, the JEGO scan also incorporates the skip scan strategy introduced in EVMamba \cite{efficientvmamba},
which scans features at intervals rather than exhaustively to reduce the sequence length.
However, EVMamba only performs forward scans, 
resulting in a global receptive field restricted to the bottom-right of the image and lacking omnidirectionality.
By arranging the starting points \raisebox{-2pt}{\includegraphics[width=10pt]{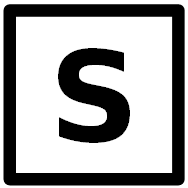}} 
and ending points \raisebox{-2pt}{\includegraphics[width=10pt]{graph/icon_end.png}} of the sequences in four directions as in \cref{properties}, 
the JEGO scan achieves a balanced receptive field and omnidirectionality.

Four sequences $\{S_i\}_{i=1}^{4}$ of length $N/4$ in four directions are generated through horizontal scan on $X^{h}$ and vertical scan on $X^{v}$.
$S_1, S_2, S_3, S_4\in \mathbb{R}^{2H_{c}W_{c}/p^{2} \times C_{1}}$ represent the sequences of right, left, up, and down scans, respectively.
These sequences are then processed independently by four $\mathrm{Mamba}$ blocks to establish long-range dependencies.
\begin{equation} \label{eq:scan}
    \begin{aligned}
    &S_{1,2} \xleftarrow{\text{horizontal scan}} X^{h}[m::p, n::p, :], \\
    &S_{3,4} \xleftarrow{\text{vertical scan}} X^{v}[m::p, n::p, :], \\
    &S_{2,4} = \mathrm{Flip}(S_{2,4}), \\
    &\{\tilde{S}_i\}_{i=1}^{4} = \mathrm{Mamba}(\{S_i\}_{i=1}^{4}), \\
    \end{aligned}
\end{equation}
where the operation $[m::p, n::p, :]$ denotes slicing the matrix for each channel, 
starting at $m$ in height and $n$ in width, while skipping every $p=2$ steps. 
The starting point \raisebox{-2pt}{\includegraphics[width=10pt]{graph/icon_start.png}} $(m, n)$ for each sequence is arranged as
\begin{equation} \label{eq:mn}
    \begin{aligned}
    &(m, n) = \left( \left\lfloor \frac{i-1}{2} \right\rfloor, (i-1) \bmod 2 \right), \quad i \in \{1, 2, 3, 4\}.
    \end{aligned}
\end{equation}

\vspace{-6mm}
\subsubsection{JEGO Merge}
\vspace{-1mm}
\label{subsec:merge}
After establishing long-range dependencies in each sequence individually, merging the sequences into a 2D feature map is crucial for visual tasks.
To address this, we propose the JEGO merge strategy to produce global omnidirectional feature maps for two images.

First, we restore the features in four sequences to their original scanned positions, 
resulting in the joint feature maps $Y^h \in \mathbb{R}^{2H_{c} \times W_{c} \times C_{1}}$ and $Y^v \in \mathbb{R}^{H_{c} \times 2W_{c} \times C_{1}}$,
\begin{equation} \label{eq:merge1}
    \begin{aligned}
    &\tilde{S}_{2,4} = \mathrm{Flip}(\tilde{S}_{2,4}), \\
    &Y^h[m::p, n::p, :] \xleftarrow{\text{restore}} \tilde{S}_{1, 2}, \\
    &Y^v[m::p, n::p, :] \xleftarrow{\text{restore}} \tilde{S}_{3, 4}. \\
    \end{aligned}
\end{equation}
Then we spatially split the joint feature maps $Y^h$ and $Y^v$ into the features of image $I_A$ and $I_B$, \eg, $Y^h \rightarrow Y^h_A, Y^h_B$.
The horizontal and vertical features are summed to obtain the merged feature maps $\tilde{F}^{c}_{A}, \tilde{F}^{c}_{B}\in\mathbb{R}^{H_{c} \times W_{c} \times C_{1}}$ as
\begin{equation} \label{eq:merge2}
    \begin{aligned}
    &\tilde{F}^{c}_{A} =Y^h_A + Y^v_A= Y_h[:H_{c}, :, :] + Y_v[:, :W_{c}, :], \\
    &\tilde{F}^{c}_{B} =Y^h_B + Y^v_B= Y_h[H_{c}:, :, :] + Y_v[:, W_{c}:, :]. \\
    \end{aligned}
\end{equation}
Although $\tilde{F}^{c}_{A}, \tilde{F}^{c}_{B}$ are omnidirectional and exhibit balanced receptive fields at the macro level, 
each individual feature still perceives unidirectional information and has local receptive fields at the micro level, 
except at the ends \raisebox{-2pt}{\includegraphics[width=10pt]{graph/icon_end.png}} of the four sequences.
Therefore, we propose using a gated convolutional unit as an aggregator for local information aggregation in $3\times3$ windows,
consolidating information from \emph{different directions and receptive fields} toward the center as
\begin{equation} \label{eq:merge3}
    \begin{aligned}
    \sigma &= \mathrm{GELU}(\mathrm{Conv_3}(\tilde{F}^{c})), \\
    \hat{F}^{c} &= \mathrm{Conv_3}(\sigma \cdot \mathrm{Conv_3}(\tilde{F}^{c})). \\
    \end{aligned}
\end{equation}
$\mathrm{Conv_3}$ denote 2D convolutions with a kernel size of $3$, 
and $\sigma$ is the gating signal that determines which information is filtered based on the input.
The aggregator ensures that $\hat{F}^{c}_{A}$ and $\hat{F}^{c}_{B}$ are \emph{global} and \emph{omnidirectional} at the micro level,
which is essential to the performance of JamMa.

\subsection{Coarse-to-Fine Matching}
\label{subsec:c2f}
\begin{figure}[t]
	\includegraphics[width=0.99\linewidth]{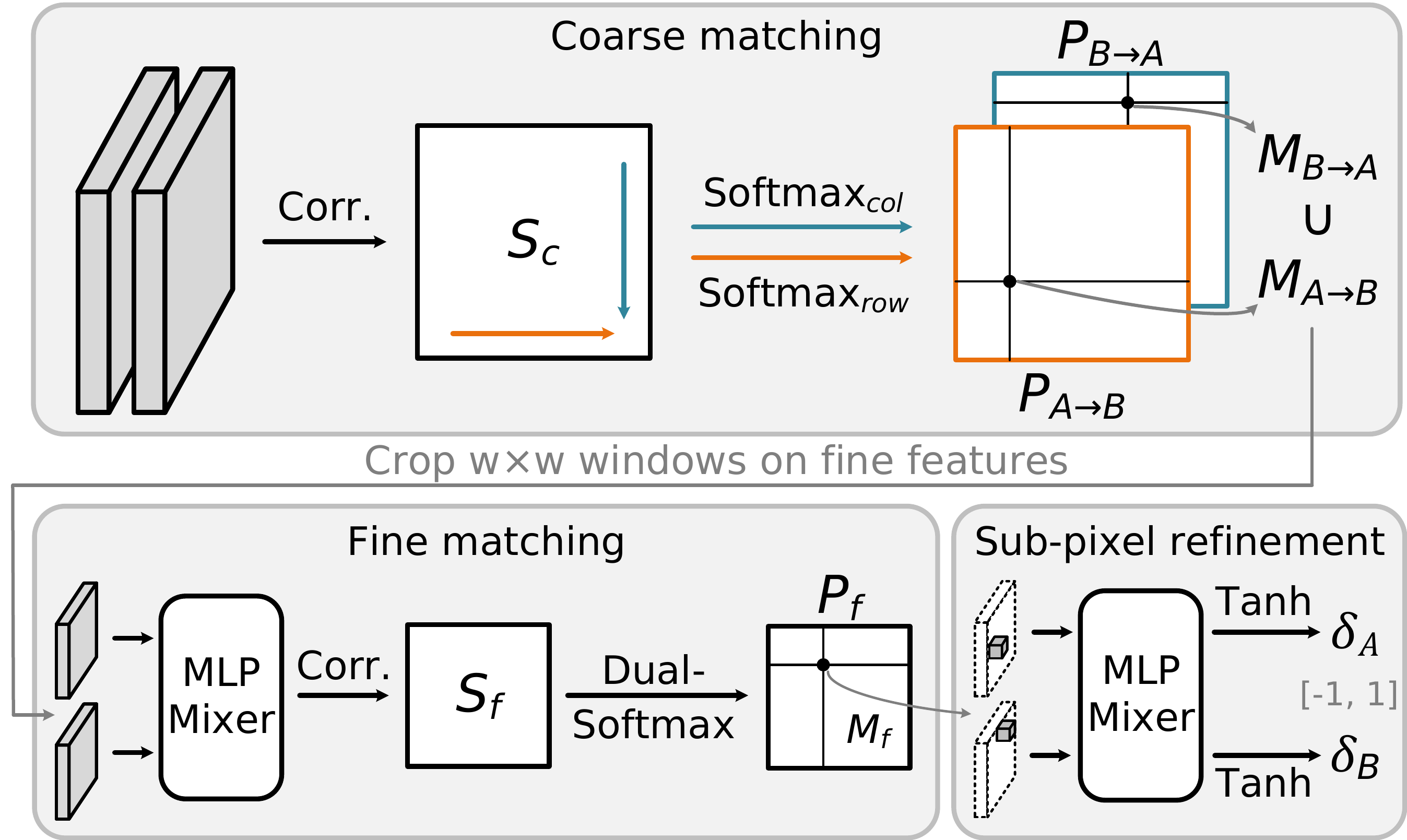}
	\centering
    \vspace{-2mm}
	 \caption{
    \textbf{Coarse-to-Fine Matching (C2F) Module.}
	 }
	 \label{c2f}
     \vspace{-6mm}
\end{figure}

Based on the aggregated coarse features $\hat{F}^{c}_{A,B}$ and the fine features $F^{f}_{A,B}$,
we adopt the coarse-to-fine matching module in XoFTR \cite{xoftr} to generate matches, as shown in \cref{c2f}.

\noindent\textbf{Coarse Matching.}
The coarse similarity matrix $S_c$ between two sets of coarse features is computed as 
\begin{equation}
    \begin{aligned}
    S_{c}(i,j)=\frac{1}{\tau}\cdot\left \langle\hat{F}^{c}_{A}(i),\hat{F}^{c}_{B}(j)\right \rangle,
    \end{aligned}
    \label{eq:inner}
\end{equation}
where $\tau$ is a temperature parameter and $\langle \cdot, \cdot \rangle$ denotes the inner product.
We then perform row $\mathrm{Softmax}_{row}$ and column $\mathrm{Softmax}_{col}$ separately on the similarity matrix to 
obtain matching probability matrices $P_{A\rightarrow B}$ and $P_{B\rightarrow A}$ as
\begin{equation}
    \begin{aligned}
    P_{A\rightarrow B} &=\mathrm{Softmax}_{row}(S_{c}), \\
    P_{B\rightarrow A} &=\mathrm{Softmax}_{col}(S_{c}).
    \end{aligned}
\end{equation}
We can assign $\hat{F}^{c}_{A}$ to $\hat{F}^{c}_{B}$ via $P_{A\rightarrow B}$ in a many-to-one fashion \cite{rcm},
and similarly assign $\hat{F}^{c}_{B}$ to $\hat{F}^{c}_{A}$ via $P_{B\rightarrow A}$,
resulting in two sets of matches $M_{A\rightarrow B}$ and $M_{B\rightarrow A}$.
Specifically, we establish $M_{A\rightarrow B}$ and $M_{B\rightarrow A}$ following two criteria: 
(1) row-maximum in $P_{A\rightarrow B}$ or column-maximum in $P_{B\rightarrow A}$, and (2) confidences greater than the threshold $\theta_{c}$.
The union of $M_{A\rightarrow B}$ and $M_{B\rightarrow A}$ is then considered as the coarse matches $M_{c}$,
which is more robust than one-to-one matches obtained through $\mathrm{Dual}$-$\mathrm{Softmax}$ operation.
\begin{equation}
    \begin{aligned}
        M_{c}~=~&\{\left(i, j\right) \mid P^{A\rightarrow B}_{i, j} = \max_k \mathcal{P}^{A\rightarrow B}_{i, k}, P^{A\rightarrow B}_{i, j} \geq \theta_{c}\} \\
        \cup~~&\{\left(i, j\right) \mid P^{B\rightarrow A}_{i, j} = \max_k \mathcal{P}^{B\rightarrow A}_{k, j}, P^{B\rightarrow A}_{i, j} \geq \theta_{c}\}.
        \label{eq:coarse-matches}
    \end{aligned}
\end{equation}

\noindent\textbf{Fine Matching.} 
For each coarse match, two $5\times5$ feature windows $\hat{F}^{f}_{A},\hat{F}^{f}_{B} \in\mathbb{R}^{M \times 25 \times C_{2}}$ are cropped from the fine features $F^{f}_{A},F^{f}_{B}$, 
where $M$ is the number of coarse matches. 
$\hat{F}^{f}_{A}$ and $\hat{F}^{f}_{B}$ are then processed by the MLP-Mixer \cite{mlpmixer} to enable mutual interaction in a lightweight manner.
Similar to \cref{eq:inner}, the fine similarity matrix $S_{f}\in\mathbb{R}^{M \times 25 \times 25}$ is computed by the inner product,
and the $\mathrm{Dual}$-$\mathrm{Softmax}$ is applied to calculate fine matching probability matrix $P_f$ as
\begin{equation}
    \begin{aligned}
    P_{f} &=\mathrm{Softmax}_{row}(S_{f}) \cdot \mathrm{Softmax}_{col}(S_{f}). \\
    \end{aligned}
\end{equation}
We further apply the mutual nearest neighbor (MNN) criteria to establish one-to-one fine matches $M_f$.

\noindent\textbf{Sub-pixel Refinement.} 
Since classification-based matches $M_f$ cannot achieve sub-pixel accuracy,
regression-based refinement is employed to further adjust the matches.
For each fine match, $F^{s}_{A},F^{s}_{B} \in\mathbb{R}^{M \times 1 \times C_{2}}$ are cropped from the feature windows,
which are then processed by the MLP-Mixer.
The $\mathrm{Tanh}$ activation function is used to compute offsets $\delta_{A},\delta_{B}\in[-1, 1]^{M \times 2}$, 
which are then added to $M_f$ for adjustment, resulting in final sub-pixel matches $M_s$.

\subsection{Supervision}
Our loss function comprises three components: the coarse matching loss $\mathcal{L}_{c}$, 
the fine matching loss $\mathcal{L}_{f}$, and the sub-pixel loss $\mathcal{L}_{s}$.
The ground-truth matching matrices $P_{c}^{gt}$ and $P_{f}^{gt}$ are generated from the camera poses and depth maps.

The coarse matching loss $\mathcal{L}_{c}$ is computed as the focal loss $\mathrm{FL}$ between $P_{c}^{gt}$ and $P_{A\rightarrow B}, P_{B\rightarrow A}$,
\begin{equation}
	\mathcal{L}_{c} = \mathrm{FL}(P_{c}^{gt}, P_{A\rightarrow B}) + \mathrm{FL}(P_{c}^{gt}, P_{B\rightarrow A}).
	\label{equ7} 
\end{equation}

The fine matching loss $\mathcal{L}_{f}$ is computed as the focal loss between $P_{f}^{gt}$ and $P_{f}$,
\begin{equation}
	\mathcal{L}_{f} = \mathrm{FL}(P_{f}^{gt}, P_{f}).
	\label{equ8} 
\end{equation}

In line with XoFTR \cite{xoftr}, we implement the symmetric epipolar distance function to compute the sub-pixel refinement loss as
\begin{equation}
    \label{eq:sym_epipolar_loss}
        \mathcal{L}_{s} = \\
        \frac{1}{\left| M_f\right|}\sum_{(x, y)} \| x^T E y \|^2 \left( \frac{1}{\|E^T x\|_{0:2}^2} + \frac{1}{\|E y\|_{0:2}^2}\right),\\
\end{equation}
where $x$ and $y$ are the homogeneous coordinates of matching points $M_s$, and $E$ denotes the ground-truth essential matrix.

\section{Experiments}
\label{sec:experiments}

\begin{table*}[t]
    \centering
    \resizebox{0.9\textwidth}{!}{
    \begin{tabular}{clcccccccccc} 
    \toprule
    \multirow{2}{*}{Category} & \multirow{2}{*}{Method} & \multirow{2}{*}{Avg. Rank $\downarrow$} & \multicolumn{3}{c}{Efficiency $\downarrow$}  & \multicolumn{3}{c}{Relative Pose Estimation $\uparrow$}   \\ 
    \cmidrule(lr){4-6}
    \cmidrule(lr){7-9}
        &      &   &Params~(M)       & FLOPs~(G)       & Time~(ms)     &AUC@5$^\circ$       &AUC@10$^\circ$       &AUC@20$^\circ$     \\ 
    \midrule
    \multirow{5}{*}{Sparse} 
    & XFeat \cite{xfeat} &7.5 &\fs 0.7  &\fs 15.7 &\fs 14.2 & 44.2 &58.2 &69.2   \\
    & SP \cite{superpoint} + SG \cite{superglue} &9.2 &13.3 &480.2 &96.9  & 57.6 &72.6 &83.5 \\
    & SP \cite{superpoint} + LG \cite{lightglue} &\rd 7.3 &13.2 & 459.9 &84.2 & 58.8 &73.6 &84.1  \\
    & DeDoDe$_B$ \cite{dedode} &10.7 &28.1 &1268.2 &189.1 &61.1 &73.8 &83.0   \\
    & DeDoDe$_G$ \cite{dedode} &8.5 &33.4 &1304.8 &329.5 &62.8 &76.3 &85.8   \\

    \hline
    \multirow{2}{*}{Dense} 
    & DKM \cite{dkm} &7.5 & 72.3 &1424.5 &554.2 &\nd 67.3 &\nd 79.7 &\nd 88.1  \\ %
    & RoMa \cite{roma} &7.5 &111.3 &2014.3 &824.9  &\fs 68.5 &\fs 80.6&\fs 88.8 \\ %

    \hline
    \multirow{7}{*}{Semi-Dense} 
    & XFeat$^\star$ \cite{xfeat} &7.5 &\nd 1.5  &\nd 48.4 &\nd 29.0 & 50.8 &66.8 &78.8   \\
    & RCM \cite{rcm} &7.5 &9.8  &363.6 &93.0 & 58.3 & 72.8 & 83.5  \\
    & LoFTR \cite{loftr} &\rd 7.3 & 11.6 &815.4 &117.5 &62.1&75.5&84.9  \\
    & MatchFormer \cite{matchformer} &8.3 &20.3  &811.1 &186.0 &62.0&75.6&84.9  \\
    & ASpanFormer \cite{aspanformer} &\rd 7.3 &15.8  &882.3 &155.7 & 62.6&76.1&85.7  \\
    & ELoFTR \cite{eloftr} &\nd 5.8 &16.0 &968.8 & 69.6 &63.7&77.0&86.4  \\
    & \name{} &\fs 3 &\rd 5.7 &\rd 202.9 &\rd 59.9 &\rd 64.1&\rd 77.4&\rd 86.5  \\

    \bottomrule
    \end{tabular}
    }
    \caption{\textbf{Results of Relative Pose Estimation on MegaDepth \cite{megadepth} Dataset.}
    The AUC of pose error at three thresholds and the overall efficiency of models are presented.
    To indicate the performance-efficiency balance, we report the average ranking across six metrics of efficiency and relative pose estimation.
    The \colorbox{first}{1st}, \colorbox{second}{2nd}, and \colorbox{third}{3rd}-best methods are highlighted.
    }
    \label{tab:exp relativepose}
    \vspace{-3mm}
    \end{table*}

\subsection{Implementation Details}\label{4.1}
The proposed method is implemented using Pytorch \cite{pytorch}.
For feature extraction, we adopt the first two stages of ConvNeXt V2-N, which contains $0.65$M parameters.
\name{} comprises $4$ Mamba blocks that process sequences in four directions.
The channel dimensions for the coarse and fine features are $C_{1} = 256$ and $C_{2} = 64$, respectively.
The coarse feature resolution is $1/8$, and the fine feature resolution is $1/2$.
The temperature parameter $\tau$ is set to $0.1$, and the coarse matching threshold $\theta_{c}$ is set to $0.2$. 
\name{} is trained on the MegaDepth dataset \cite{megadepth} for $30$ epochs with a batch size of $2$ using the AdamW optimizer \cite{adamw}, and it is not fine-tuned for any other tasks.
During training, images are resized and padded to a size of $832 \times 832$. 
The initial learning rate is set to $0.0002$, and a cosine decay learning rate scheduler with $1$ epoch of linear warm-up is employed.
Network training takes $\sim$50 hours on a single NVIDIA 4090 GPU, and all evaluations are also conducted on this GPU.

\subsection{Relative Pose Estimation}
\label{sec:mega}
\textbf{Dataset.}
We evaluate matchers on the MegaDepth dataset \cite{megadepth} for pose estimation. 
Test images are resized and padded to $832$$\times$$832$, the same as the training size.
As suggested in \cite{lightglue}, we use LO-RANSAC \cite{poselib} to estimate essential matrix for all methods,
as it is more robust than vanilla RANSAC, enhancing evaluation reliability.
The detailed evaluation setup of each method is reported in the Appendix.

\noindent\textbf{Metric.}
The area under the cumulative curve (AUC) of the pose error at thresholds $(5^\circ, 10^\circ, 20^\circ)$ are reported.
We also report three efficiency metrics including parameters, FLOPs, and runtime, along with the average ranking across six metrics of efficiency and performance.

\noindent\textbf{Results.}
With 14 evaluated methods, an average ranking of $7.5$ is defined as the expected performance-efficiency balance, 
\eg, the best performance with the worst efficiency. 
As presented in \cref{tab:exp relativepose}, JamMa achieves an average ranking of $3.5$, leading the expected balance and other matchers by a large margin.
In terms of performance, \name{} demonstrates the best pose estimation results among evaluated semi-dense and sparse matchers, surpassed only by the dense matchers DKM and RoMa that prioritize accuracy.
In terms of efficiency, \name{} exceeds all evaluated matchers except XFeat and XFeat$^\star$.
JamMa has notably fewer parameters than attention-based matchers, making it ultra-lightweight.
\begin{table}[t]
    \centering
    \resizebox{0.9\columnwidth}{!}{
    \begin{tabular}{clccc} 
    \toprule
    \multirow{2}{*}{Category} & \multirow{2}{*}{Method}         & \multicolumn{3}{c}{Homography est. AUC} \\ 
    \cmidrule(lr){3-5}
        &              & @3px       & @5px       & @10px \\ 
    \midrule
    \multirow{5}{*}{Sparse} 
    & SP \cite{superpoint} + NN & 41.6 & 55.8 & 71.7  \\
    & R2D2 \cite{r2d2} + NN & 50.6 & 63.9 &76.8\\
    & DISK \cite{disk} + NN & 52.3 & 64.9 & 78.9\\
    & SP \cite{superpoint} + SG \cite{superglue} & 53.9 &68.3 &81.7 \\
    & SP \cite{superpoint} + LG \cite{lightglue} & 54.2 &68.3 & 81.5 \\
    \hline
    \multirow{5}{*}{Semi-Dense} 
    & DRC-Net \cite{dual} & 50.6 & 56.2 & 68.3 \\
    & LoFTR \cite{loftr} & 65.9 & 75.6 & 84.6\\
    & ASpanFormer \cite{aspanformer} &\nd 67.4 &\nd 76.9 &\fs 85.6\\
    & ELoFTR \cite{eloftr} &\rd 66.5&\cellcolor{third}76.4&\nd 85.5 \\
    & \name{} &\fs 68.1&\fs 77.0&\rd 85.4 \\
    \bottomrule
    \end{tabular}
    }
    \vspace{-2mm}
    \caption{\textbf{Results of Homography Estimation on HPatches \cite{hpatches} Dataset.}
    The AUC of reprojection error of corner points at different thresholds are reported.
    }
    \label{tab:exp hpatches}
    \vspace{-5mm}
\end{table}

\subsection{Homography Estimation}
\textbf{Dataset.}
The HPatches dataset \cite{hpatches} contains planar scenes for evaluating homography estimation.
We assess the performance of our method across 108 HPatches sequences, 
which comprise 52 sequences with illumination variations and 56 sequences with viewpoint changes.

\noindent\textbf{Metric.}
Following the evaluation protocol proposed in LoFTR \cite{loftr},
we compute the mean reprojection error of corner points, and report the AUC
of the corner error up to threshold values of $3$, $5$, and $10$ pixels, respectively.
All images are resized so that their smaller dimension is equal to $480$ pixels.
Vanilla RANSAC \cite{ransac} is applied to estimate the homography for all methods.
The top $1000$ matches of semi-dense methods are selected for the sake of fairness.

\noindent\textbf{Results.}
As shown in \cref{tab:exp hpatches}, \name{} achieves competitive performance in homography estimation.
Comparing the state-of-the-art matcher ASpanFormer, \name{} lags behind by $0.2\%$ at threshold $10$ but leads by $0.7\%$ at threshold $3$.
It is worth noting that \name{} is much more lightweight than ASpanFormer, with only $36\%$ of its parameters.
We attribute this favorable performance-efficiency balance to the global omnidirectional representation resulting from the JEGO strategy and the Mamba with linear complexity.

\begin{figure*}[t]
	\includegraphics[width=0.9\linewidth]{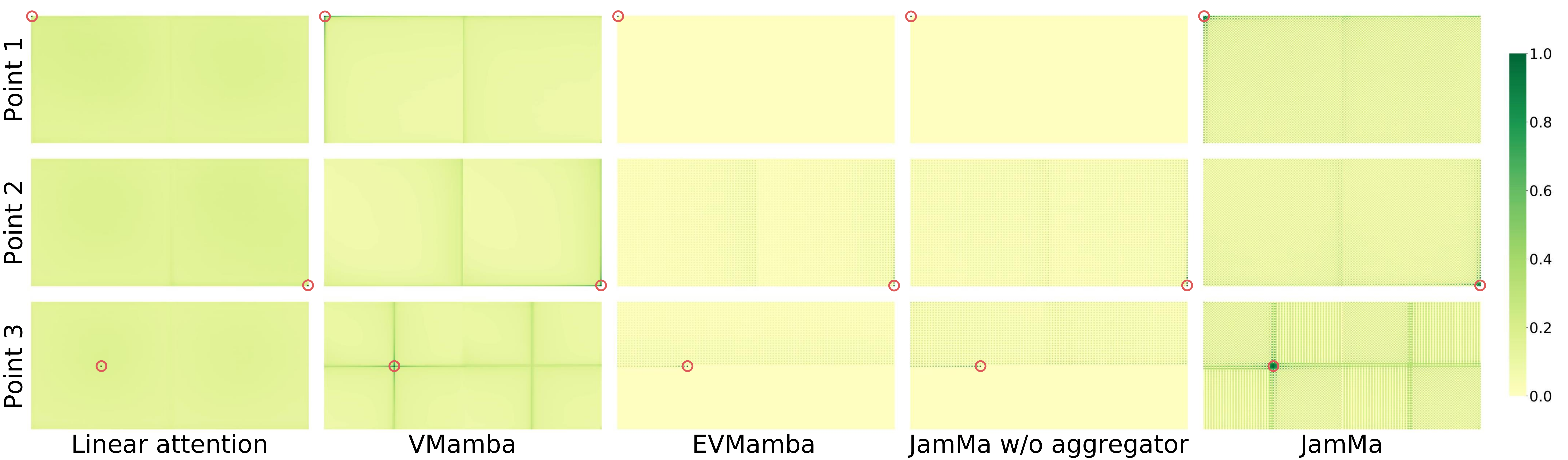}
	\centering
	\vspace{-3mm}
	 \caption{
    \textbf{Effective Receptive Field (ERF).}
	Red circles denote query points, and the ERFs of query points in the two images are concatenated horizontally.
	\name{} achieves a global and omnidirectional receptive field, a capability made possible through the aggregator.
	 }
	 \label{erf}
	 \vspace{-5mm}
\end{figure*}

\begin{table}[t]
    \centering
    \resizebox{0.9\columnwidth}{!}{
    \begin{tabular}{lcccc} 
    \toprule
    \multirow{2}{*}{Method}     & Time$^*$   &\multicolumn{3}{c}{Pose est. AUC} \\ 
    \cmidrule(lr){3-5}
    &(ms) &{~@$5^\circ$~}  &{@$10^\circ$}  &{@$20^\circ$} \\ 
    \midrule
    \name{} &\rd 3.2 & \fs 64.5 &\fs 77.3 &\fs 86.3  \\
    (1) Replace JS to SS &\rd 3.2 & 62.2&74.7&83.7 \\
    (2) w/o Aggregator  &\nd 3.0 &62.3 & 75.1 & 84.3 \\
    (3) w/ EVMamba scan &\nd 3.0 & 61.9 & 74.8 & 84.1  \\ 
    (4) w/ VMamba scan & 9.7 &\rd 64.1 &\nd 77.1 &\nd 86.2  \\ 
    (5) w/ van. attention & \multicolumn{4}{c}{OOM} \\ 
    (6) w/ lin. attention &24.3 &\nd 64.2 &\rd 77.0 &\rd 86.1  \\
    (7) w/o interaction &\fs 0 &60.1 &73.0 &82.6  \\
    \bottomrule
    \end{tabular}
    }
    \vspace{-2mm}
    \caption{\textbf{Ablation Study.}
    SS and JS denote sequential scan and joint scan, respectively.
    The superscript $^*$ indicates that the time of the coarse layer, including scanning and merging, is reported.
    }
    \label{tab:exp ablation}
    \vspace{-5mm}
\end{table}

\subsection{Ablation Study}
We conduct qualitative and quantitative ablation studies on the MegaDepth dataset for a comprehensive understanding of \name{}.
Models in \cref{tab:exp ablation} are trained for $15$ epochs at $544$ resolution and tested with the same settings as in \cref{sec:mega}.

\noindent\textbf{Joint Scan.}
As shown in \cref{tab:exp ablation}(1),
replacing the joint scan with the sequential scan leads to a significant performance drop of ($-2.3\%, -2.6\%, -2.6\%$),
which suggests that the joint scan is more effective for feature matching, 
as it promotes high-frequency mutual interactions between the two images.
Note that VMamba and EVMamba inherently lack mutual interaction capabilities, 
and thus, we introduce the joint scan to both methods for ablation studies (3) and (4).

\noindent\textbf{Aggregator.}
We present the effective receptive fields (ERF) in \cref{erf}, 
where the coarse features of \name{} achieve global receptive fields and perceive information from all directions.
Without the aggregator, \name{} loses the global receptive field and omnidirectionality as previously analyzed.
The results in \cref{tab:exp ablation}(2) indicate a severe performance decline of ($-2.2\%, -2.2\%, -2\%$) without the aggregator.
More ablation studies of the aggregator are reported in the Appendix.
Additionally, the comparison between (2) and (3) confirms that the four-directional scan outperforms the bi-directional scan in EVMamba, even in the absence of the aggregator.

\begin{figure}[t]
	\includegraphics[width=0.9\linewidth]{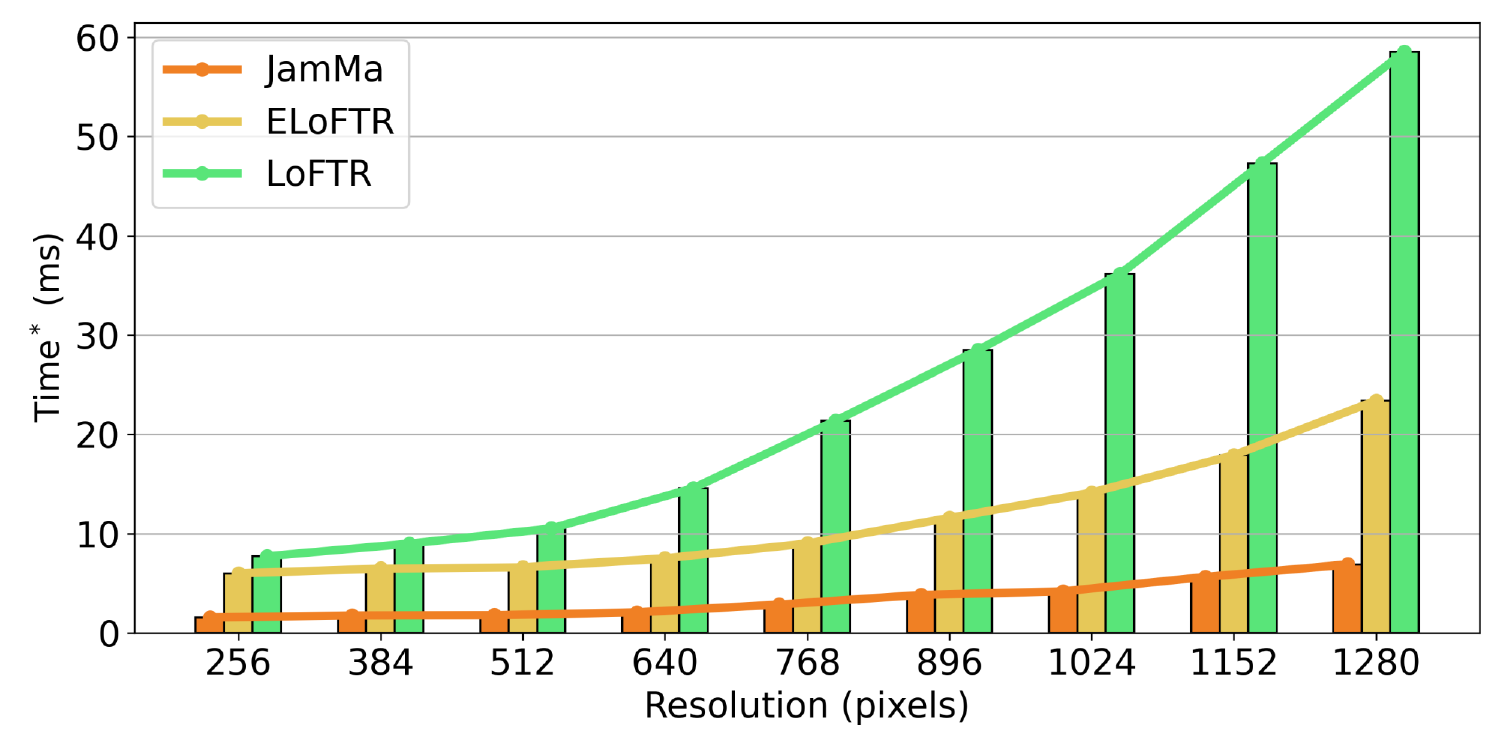}
	\centering
	\vspace{-4mm}
	 \caption{
		\textbf{Comparison of Coarse Layer Processing Times.}
	 }
	 \label{coarse_time}
	 \vspace{-5mm}
\end{figure}

\noindent\textbf{Comparison with Mamba-based Methods.}
As shown in \cref{tab:exp ablation}(3)(4),
we evaluate the JEGO strategy against those proposed in EVMamba and VMamba.
Compared to EVMamba, \name{} exhibits a substantial performance improvement of ($+2.6\%, +2.5\%, +2.2\%$).
While VMamba achieves a global receptive field as shown in \cref{erf},
this comes at the cost of $4\times$ total sequence length. 
Leveraging the JEGO strategy, 
JamMa layer achieves $3\times$ speedup over the VMamba layer while delivering superior performance,
resulting in a $9.5\%$ overall speed improvement ($68.3$ms vs. $61.8$ms).

\noindent\textbf{Comparison with Attention-based Methods.}
In the case of (5) vanilla attention, the out-of-memory (OOM) error occurs in 24GB VRAM.
Comparing (6) linear attention, 
\name{} achieves a performance boost of ($+0.3\%, +0.3\%, +0.2\%$) while being $7.6\times$ faster.
JamMa minimizes the bottleneck of most matchers, \emph{i.e.}, attention-based interaction, 
to nearly negligible levels, significantly enhancing overall efficiency. 
The total runtimes of JamMa and (6) are $82.9$ms vs. $61.8$ms (a $25\%$ total speed gain),
and $4$ JamMa layers and $4$ linear attention layers contain parameters of $1.8$M vs. $5.3$M (a $66\%$ parameter reduction).
\cref{coarse_time} further shows the processing time of coarse layers at various resolutions.

\begin{figure*}[t]
	\includegraphics[width=0.85\linewidth]{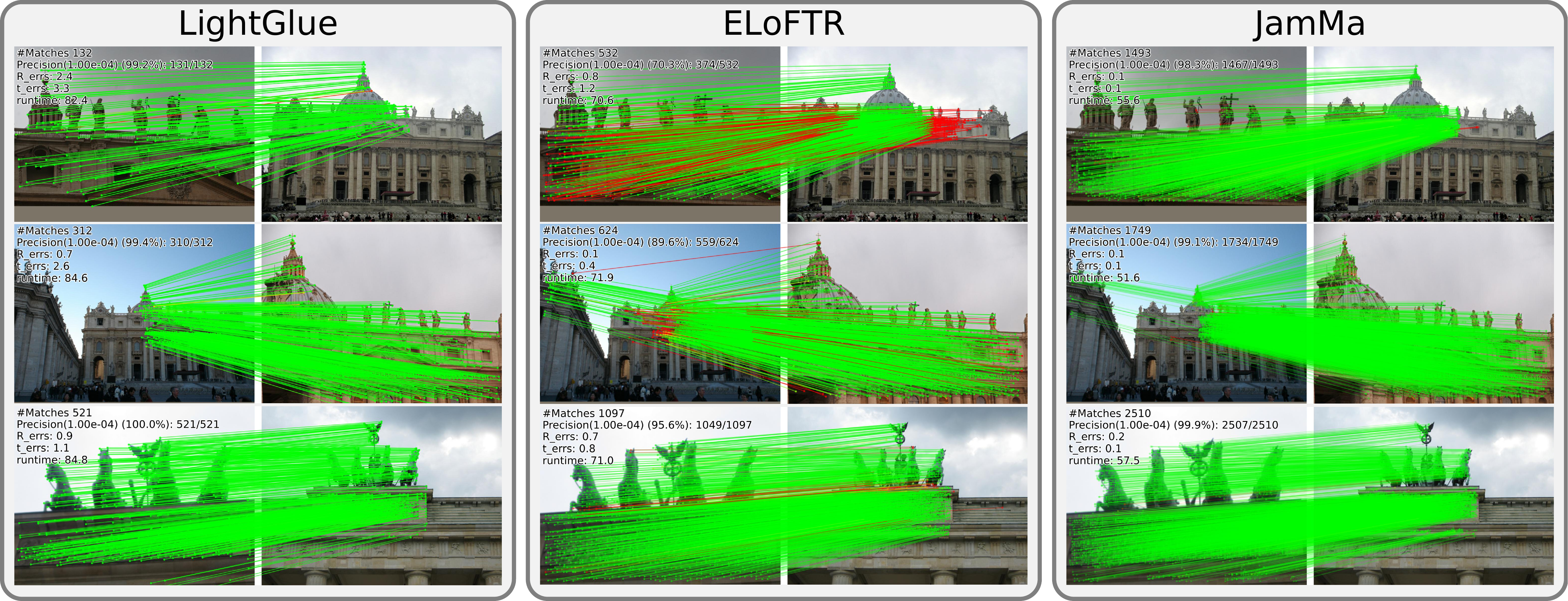}
	\centering
	\vspace{-2mm}
	 \caption{
    \textbf{Comparison of Qualitative Results.} 
	The reported metrics include precision with an epipolar error threshold of $1 \times 10^{-4}$, rotation and translation errors in pose estimation, and runtime. 
	\name{} consistently delivers more robust matching results with shorter runtime.
	}
	 \label{viz}
	 \vspace{-4mm}
\end{figure*}

\begin{table}[t]
	\centering
  \resizebox{0.85\linewidth}{!}{
    \begin{tabular}{l c c c c c c}
			\toprule
			\multirow{2}{*}{Method} &Params &Time  &\multicolumn{3}{c}{Pose est. AUC} \\
			\cmidrule(lr){4-6}
			&(M) &(ms) 	&{@$5^\circ$}  &{@$10^\circ$}  &{@$20^\circ$} \\
			\midrule
      JamMa     &5.7 &61.8 &64.5  &77.3 &86.3  \\
      \midrule
      w/ XFeat &5.7 &54.6 &61.1 &74.6 &84.1  \\
      w/ SuperPoint &6.0 &67.5 &61.8 &75.1 &84.8  \\
      w/ ResNet &8.8 &72.0 &63.8 &76.8 &85.9   \\
      \bottomrule
    \end{tabular}
  }
  \vspace{-3mm}
  \caption{\textbf{Ablation Studies on Encoder and Interaction.} 
  }
  \vspace{-3mm}
	\label{tab:encoder}
\end{table}

\noindent\textbf{Encoders.}
The variants of JamMa, employing XFeat, SuperPoint, and ResNet as encoders, are shown in \cref{tab:encoder}.
JamMa delivers competitive results with different encoders, 
with ConvNeXt offering the best performance-efficiency balance.
\section{Discussion}
\label{sec:diss}

\begin{figure}[t]
	\includegraphics[width=0.99\linewidth]{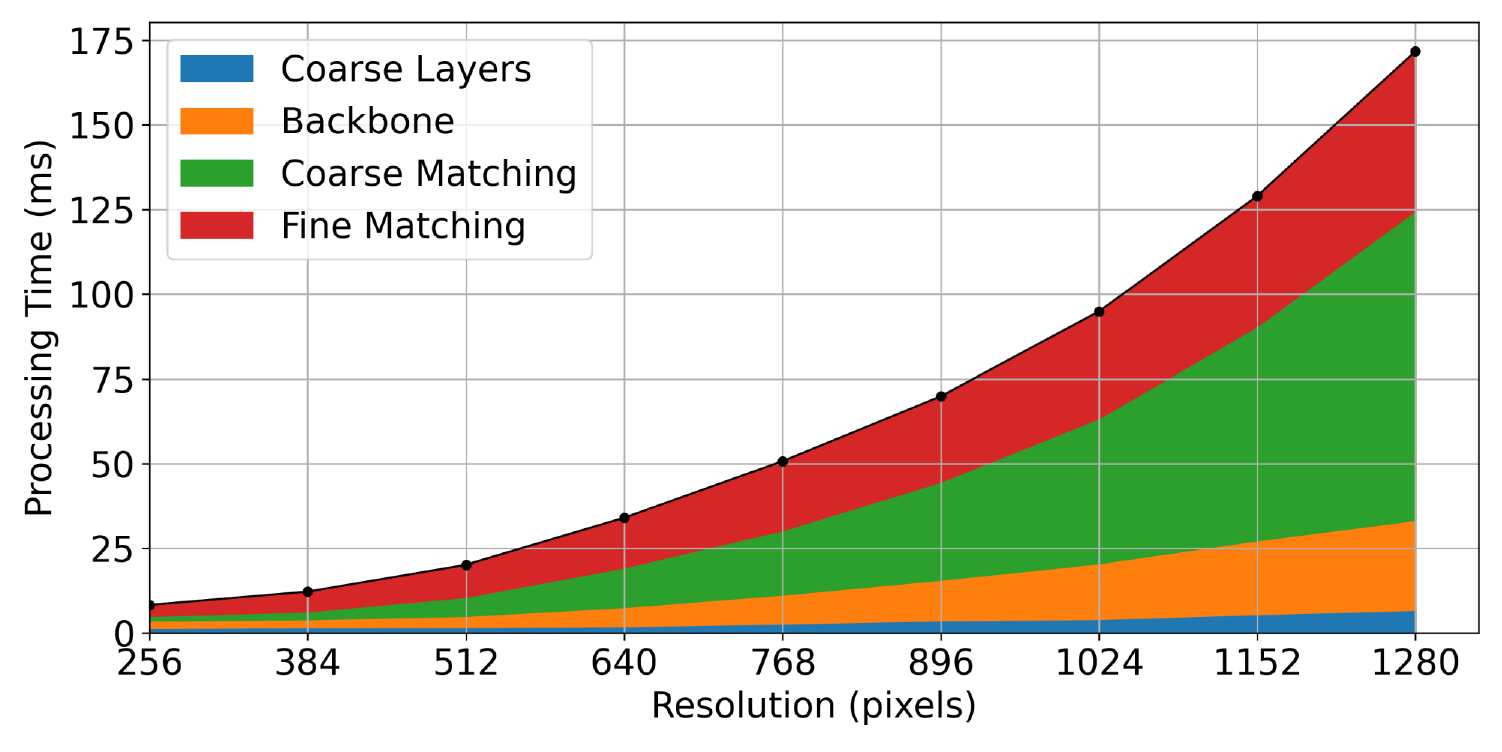}
	\centering
	\vspace{-3mm}
	 \caption{
		\textbf{Component Analysis of Runtime.}
	 }
	 \label{stage_time}
	 \vspace{-6mm}
\end{figure}

As shown in \cref{stage_time}, we conduct a component analysis on the runtime of JamMa for further acceleration.
Our findings indicate that, when processing coarse features with Joint Mamba, the coarse layers are no longer the efficiency bottleneck as observed in attention-based methods.
Instead, coarse matching using Softmax on the large similarity matrix has emerged as new bottleneck that require optimization.
We explore the Softmax-free coarse matching proposed in \cite{eloftr} and observe a $10$ms speedup without a performance drop. 
Note that this optimization is excluded in \cref{sec:experiments} for fairness.

As shown in \cref{tab:ScanNet_result},
we conduct zero-shot pose estimation experiments in the ScanNet \cite{scannet} dataset.
While \name{} achieves competitive performance in outdoor scenes, 
this does not fully transfer to indoor scenes, 
which we attribute to its reduced number of parameters (less than $50\%$ of those in compared methods), potentially limiting the generalizability.
Note that \name{} still strikes a reasonable performance-efficiency balance compared to other methods.

\begin{table}[t]
	\centering
	\resizebox{0.9\linewidth}{!}{
		\begin{tabular}{l c c c c c c}
			\toprule
			\multirow{2}{*}{Method} & {Avg.} &Params &Time  &\multicolumn{3}{c}{Pose est. AUC} \\
			\cmidrule(lr){5-7}
			&Rank &(M) &(ms) 	&{@$5^\circ$}  &{@$10^\circ$}  &{@$20^\circ$} \\
			\midrule
			SP~\cite{superpoint}+SG~\cite{superglue} &\rd 3.6 &13.3 &\rd 35.8 &\rd 16.5 &\rd 32.2 &48.8\\
			SP~\cite{superpoint}+LG~\cite{lightglue} &\nd 2.6 &\rd 13.2 &\fs 22.9 &15.9 &\rd 32.2 &\rd 49.7\\
            LoFTR \cite{loftr} &\fs 1.8 &\nd 11.6 &49.2 &\fs 20.7&\fs 37.7&\fs 53.4\\
            \name{} &\fs 1.8 &\fs 5.7 &\nd 26.6 &\nd 18.5&\nd 35.6&\nd 51.3\\
			\bottomrule
		\end{tabular}
	}
	\vspace{-2mm}
    \caption{\textbf{Relative Pose Estimation on ScanNet \cite{scannet} Dataset.}}
	\label{tab:ScanNet_result}
	\vspace{-6mm}
\end{table}
\section{Conclusion}
\label{sec:conclusion}
This paper presents \name{}, an ultra-lightweight Mamba-based feature matcher with competitive performance.
We explore various scan-merge strategies for image pairs, 
a topic not previously investigated, as existing visual Mamba models focus on single-image tasks.
Firstly, we find that joint scan with high-frequency mutual interaction substantially outperforms sequential scan, which resembles self- and cross-attention.
Secondly, a four-directional scan is combined with an aggregator to generate global and omnidirectional features, which play a crucial role in performance.
Thirdly, Mamba coupled with skip scan makes \name{} extremely efficient and even surpasses most attention-based sparse methods.
Experiments demonstrate that \name{} achieves a remarkable balance between performance and efficiency.

{
    \small
    \bibliographystyle{ieeenat_fullname}
    \bibliography{main}
}


\clearpage
\newcommand{\AppendixPrefix}{S}
\renewcommand{\thesection}{\Alph{section}}
\renewcommand{\thefigure}{\AppendixPrefix\arabic{figure}}
\renewcommand{\thetable}{\AppendixPrefix\arabic{table}}
\renewcommand{\theequation}{\AppendixPrefix\arabic{equation}}
\renewcommand{\thealgorithm}{\AppendixPrefix\arabic{algorithm}}
\setcounter{section}{0}
\setcounter{figure}{0}
\setcounter{table}{0}
\setcounter{equation}{0}
\setcounter{page}{1}

\maketitlesupplementary

The supplementary material for JamMa is organized as follows:
\cref{sec:setup_supp} reports the evaluation setups of pose estimation experiments in the MegaDepth dataset.
\cref{sec:details_supp} provides additional details on the Mamba block and MLP-Mixer employed in JamMa.
\cref{sec:experiments_supp} presents further qualitative and quantitative experiments, along with discussions.

\begin{table}[th]
    \centering
    \resizebox{0.99\linewidth}{!}{
    \begin{tabular}{cccccccccccc} 
    \toprule
    \multirow{1}{*}{Category} &Method  &Image &Keypoint  &Match\\ 
    \midrule
    \multirow{5}{*}{Sparse} 
    & XFeat & 1600 &4096 &- \\ 
    & SP + SG & 1600 &2048 &- \\ 
    & SP + LG & 1600 &2048 &-\\ 
    & DeDoDe$_B$ &784 &10000 &-\\ 
    & DeDoDe$_G$ &784 &10000 &-\\ 
    \hline
    \multirow{1}{*}{Dense} 
    & All &672 &- &5000\\ %
    \hline
    \multirow{1}{*}{Semi-Dense} 
    & All &832&-&-\\ 
    \bottomrule
    \end{tabular}
    }
    \vspace{-2mm}
    \caption{\textbf{Evaluation Setups in the MegaDepth \cite{megadepth} Dataset.}
    }
    \vspace{-3mm}
    \label{setup}
\end{table}

\section{MegaDepth Setups}
\label{sec:setup_supp}
As shown in \cref{setup}, we follow the evaluation setup specified for each method to optimize their performance.
For XFeat, images are resized such that the larger dimension is $1600$ pixels, with $4096$ keypoints extracted per image.
For SuperGlue and LightGlue, the larger dimension is similarly resized to $1600$ pixels, and $2048$ SuperPoint keypoints are extracted per image.
For DeDoDe, images are resized to  $784\times 784$, with $10000$ keypoints extracted per image.
For the dense methods DKM and RoMa, images are resized to $674\times 674$, and $5000$ balanced matches are sampled using the KDE-based method introduced by DKM.
For all semi-dense methods \cite{loftr,matchformer,aspanformer,eloftr}, images are resized and padded to $832 \times 832$. 
In the efficiency evaluation, we report the parameters, FLOPs and runtime of the full sparse matching pipelines, including detection, description and matching.
All methods utilize LO-RANSAC with an inlier threshold of $0.5$ for pose estimation.

\section{Details}
\label{sec:details_supp}
\subsection{Mamba Block}
\begin{figure}[t]
	\includegraphics[width=0.99\linewidth]{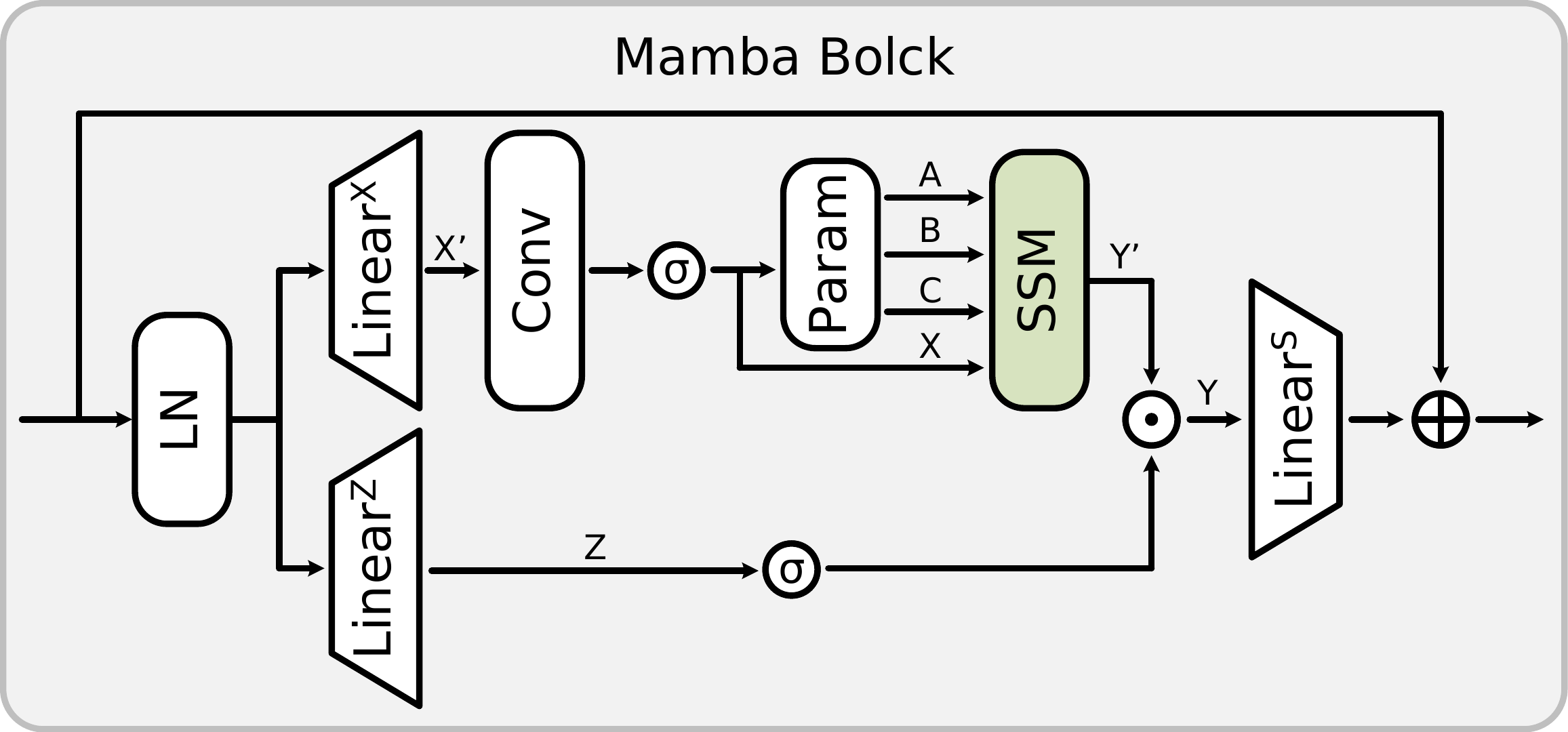}
	\centering
	\vspace{-2mm}
	 \caption{
    \textbf{Mamba Block.}
	}
	 \label{mamba_block}
\end{figure}

Details of the Mamba block \cite{mamba} are provided in \cref{mamba_block} and Alg. \ref{alg:block}.
$\mathtt{B}$ and $\mathtt{N}$ denote the batch size and sequence length, respectively.
$\mathtt{C_1}$ denotes the coarse feature dimension, which is $256$.
The SSM dimension $\mathtt{C_s}$ is set to $16$, and the expanded state dimension $\mathtt{C_e}$ is set to $512$.
The input sequence $\mathbf{S}$ is first normalized by the layer normalization $\mathrm{LN}$
and then linearly projected to $\mathbf{X}'$ and $\mathbf{Z}$, both with a dimension size of $\mathtt{C_e}$.
A 1D convolution followed by the $\mathrm{SiLU}$ nonlinearity is applied to $\mathbf{X}'$, 
producing $\mathbf{X}$, which is then linearly projected to $\mathbf{B}'$, $\mathbf{C}$, and $\mathbf{\Delta}$.
$\mathbf{\Delta}$ is used to discretize $\mathbf{A}'$ and $\mathbf{B}'$, resulting in $\mathbf{A}$ and $\mathbf{B}$.
The state-space model (SSM) computes $\mathbf{Y}'$, which is then gated by $\mathbf{Z}$ to generate $\mathbf{Y}$.
The output sequence $\tilde{\mathbf{S}}$ is obtained through the residual connection of $\mathbf{Y}$ and $\mathbf{S}$.

\begin{algorithm}[t]
    \caption{Mamba Block}
    \label{alg:block}
    \small
    \begin{algorithmic}[1]
    \REQUIRE{input sequence $\mathbf{S}$ : \textcolor{shapecolor}{$(\mathtt{B}, \mathtt{N}, \mathtt{C_1})$}}
    \ENSURE{output sequence $\tilde{\mathbf{S}}$ : \textcolor{shapecolor}{$(\mathtt{B}, \mathtt{N}, \mathtt{C_1})$}}
    \STATE $\mathbf{S}'$ : \textcolor{shapecolor}{$(\mathtt{B}, \mathtt{N}, \mathtt{C_1})$} $\leftarrow$ $\mathbf{LayerNorm}(\mathbf{S})$
    \STATE $\mathbf{X}'$ : \textcolor{shapecolor}{$(\mathtt{B}, \mathtt{N}, \mathtt{C_e})$} $\leftarrow$ $\mathbf{Linear}^\mathbf{X}(\mathbf{S}')$
    \STATE $\mathbf{Z}$ : \textcolor{shapecolor}{$(\mathtt{B}, \mathtt{N}, \mathtt{C_e})$} $\leftarrow$ $\mathbf{Linear}^\mathbf{Z}(\mathbf{S}')$
    \STATE $\mathbf{X}$ : \textcolor{shapecolor}{$(\mathtt{B}, \mathtt{N}, \mathtt{C_e})$} $\leftarrow$ $\mathbf{SiLU}(\mathbf{Conv}(\mathbf{X}'))$
    \STATE \textcolor{gray}{\text{/* compute SSM parameters, ``Param'' in Fig. S1 */}}
    \STATE $\mathbf{P}^{\mathbf{\Delta}}$ : \textcolor{shapecolor}{$(\mathtt{B}, \mathtt{N}, \mathtt{C_e})$} $\leftarrow$ $\mathbf{Parameter}$ 
    \STATE $\mathbf{\Delta}$ : \textcolor{shapecolor}{$(\mathtt{B}, \mathtt{N}, \mathtt{C_e})$} $\leftarrow$ $\mathbf{Softplus}(\mathbf{Linear}^{\mathbf{\Delta}}(\mathbf{X}) + \mathbf{P}^{\mathbf{\Delta}})$
    \STATE $\mathbf{A}'$ : \textcolor{shapecolor}{$(\mathtt{C_e}, \mathtt{C_s})$} $\leftarrow$ $\mathbf{Parameter}$ 
    \STATE $\mathbf{B}'$ : \textcolor{shapecolor}{$(\mathtt{B}, \mathtt{N}, \mathtt{C_s})$} $\leftarrow$ $\mathbf{Linear}^{\mathbf{B}}(\mathbf{X})$
    \STATE $\mathbf{A,B}$ : \textcolor{shapecolor}{$(\mathtt{B}, \mathtt{N}, \mathtt{C_e}, \mathtt{C_s})$} $\leftarrow$ $\mathbf{Discretize}(\mathbf{A}',\mathbf{B}',\mathbf{\Delta})$
    \STATE $\mathbf{C}$ : \textcolor{shapecolor}{$(\mathtt{B}, \mathtt{N}, \mathtt{C_s})$} $\leftarrow$ $\mathbf{Linear}^{\mathbf{C}}(\mathbf{X})$
    \STATE \textcolor{gray}{\text{/* SSM recurrent*/}}
    \STATE $h$ : \textcolor{shapecolor}{$(\mathtt{B}, \mathtt{C_e}, \mathtt{C_s})$} $\leftarrow$ zeros \textcolor{shapecolor}{$(\mathtt{B}, \mathtt{C_e}, \mathtt{C_s})$}
    \STATE $\mathbf{Y}$ : \textcolor{shapecolor}{$(\mathtt{B}, \mathtt{N}, \mathtt{C_e})$} $\leftarrow$ zeros \textcolor{shapecolor}{$(\mathtt{B}, \mathtt{N}, \mathtt{C_e})$}
    \FOR{$i$ in \{0, ..., N-1\}}
    \STATE $h$ = $\mathbf{A}[:,i,:,:]h + \mathbf{B}[:,i,:,:]\mathbf{X}[:,i,:,\textcolor{shapecolor}{\mathtt{None}}] $
    \STATE $\mathbf{Y}'[:,i,:]$ = $h\mathbf{C}[:,i,:]$
    \ENDFOR
    \STATE \textcolor{gray}{\text{/* get gated $\mathbf{Y}$ */}}
    \STATE $\mathbf{Y}$ : \textcolor{shapecolor}{$(\mathtt{B}, \mathtt{N}, \mathtt{C_e})$} $\leftarrow$ $\mathbf{Y}' \bigodot \mathbf{SiLU}(\mathbf{Z}) $
    \STATE \textcolor{gray}{\text{/* residual connection */}}
    \STATE $\tilde{\mathbf{S}}$ : \textcolor{shapecolor}{$(\mathtt{B}, \mathtt{N}, \mathtt{C_1})$} $\leftarrow$ $\mathbf{Linear}^\mathbf{S}(\mathbf{Y}) + \mathbf{S}$
    \STATE Return: $\tilde{\mathbf{S}}$ 
    \end{algorithmic}
\end{algorithm}

Note that the $\mathbf{for}$ loop in Alg. \ref{alg:block}, \ie, SSM recurrent, can be computed once by a global convolution as
\begin{align}
    \begin{split}
        \mathbf{K} &= (\mathbf{C}\mathbf{B},\mathbf{C}\mathbf{A}\mathbf{B}, ..., \mathbf{C}\mathbf{A}^{N-1}\mathbf{B}),\\
        \mathbf{Y}' &= \mathbf{X} \circledast \mathbf{K}, 
    \end{split}
\end{align}
where $\circledast$ denotes the convolution operation.

\subsection{MLP-Mixer}
The MLP-Mixer \cite{mlpmixer} is a purely MLP-based network that first performs spatial mixing using token-wise $\mathrm{MLP}_s$, 
followed by channel mixing using channel-wise $\mathrm{MLP}_c$.
\begin{equation} \label{eq:mlp}
    \begin{aligned}
    &F_{mid} = F_{in} + \mathrm{MLP}_s(F_{in}), \\
	&F_{out} = F_{mid} + \mathrm{MLP}_c(F_{mid}). \\
    \end{aligned}
\end{equation}

In fine matching module, two $5\times5$ fine feature windows $\hat{F}^{f}_{A},\hat{F}^{f}_{B} \in\mathbb{R}^{M \times 25 \times C_{2}}$ are spatially concatenated
to form $\hat{F}^{f}\in\mathbb{R}^{M \times 50 \times C_{2}}$, which is processed by a MLP-Mixer.

In sub-pixel refinement module, two fine features $F^{s}_{A},F^{s}_{B} \in\mathbb{R}^{M \times 1 \times C_{2}}$ are concatenated along the channel dimension,
resulting in $F^s\in\mathbb{R}^{M \times 1 \times 2C_{2}}$.
The $\mathrm{MLP}$ and $\mathrm{Tanh}$ activation are then employed to regress the offsets $\delta^x_A, \delta^y_A, \delta^x_B, \delta^y_B$ of the matching points in images $I_A$ and $I_B$.

\begin{figure*}[t]
	\includegraphics[width=0.9\linewidth]{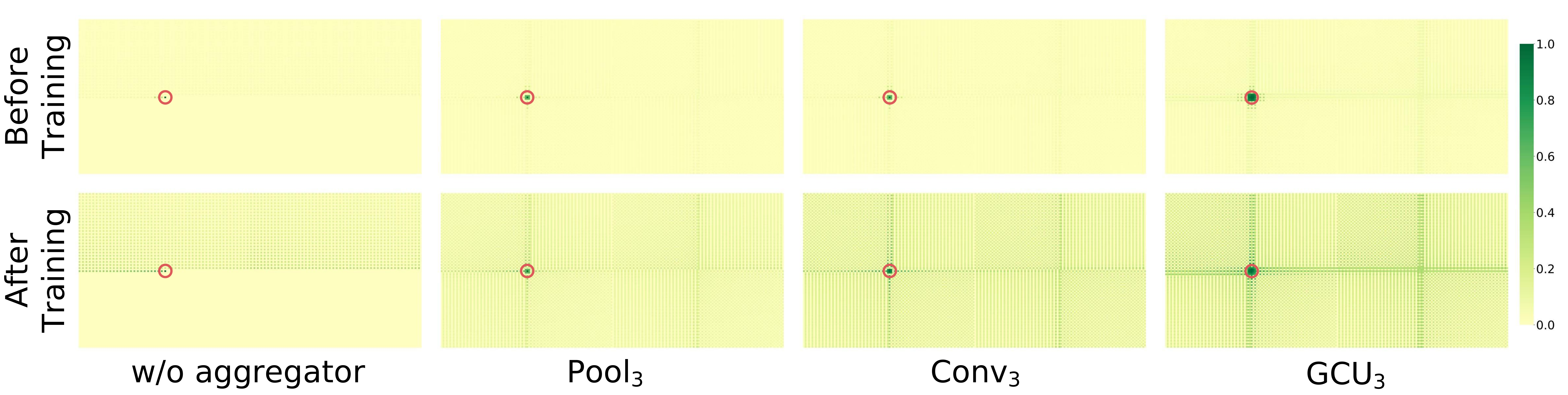}
	\centering
	\vspace{-2mm}
	 \caption{
    \textbf{Effective Receptive Field with Different Aggregators.}
	All three aggregators expand the local receptive field to a receptive field spread over the image pair.
	}
	 \label{erf_agg}
	 \vspace{-3mm}
\end{figure*}

\begin{figure}[t]
	\includegraphics[width=0.9\linewidth]{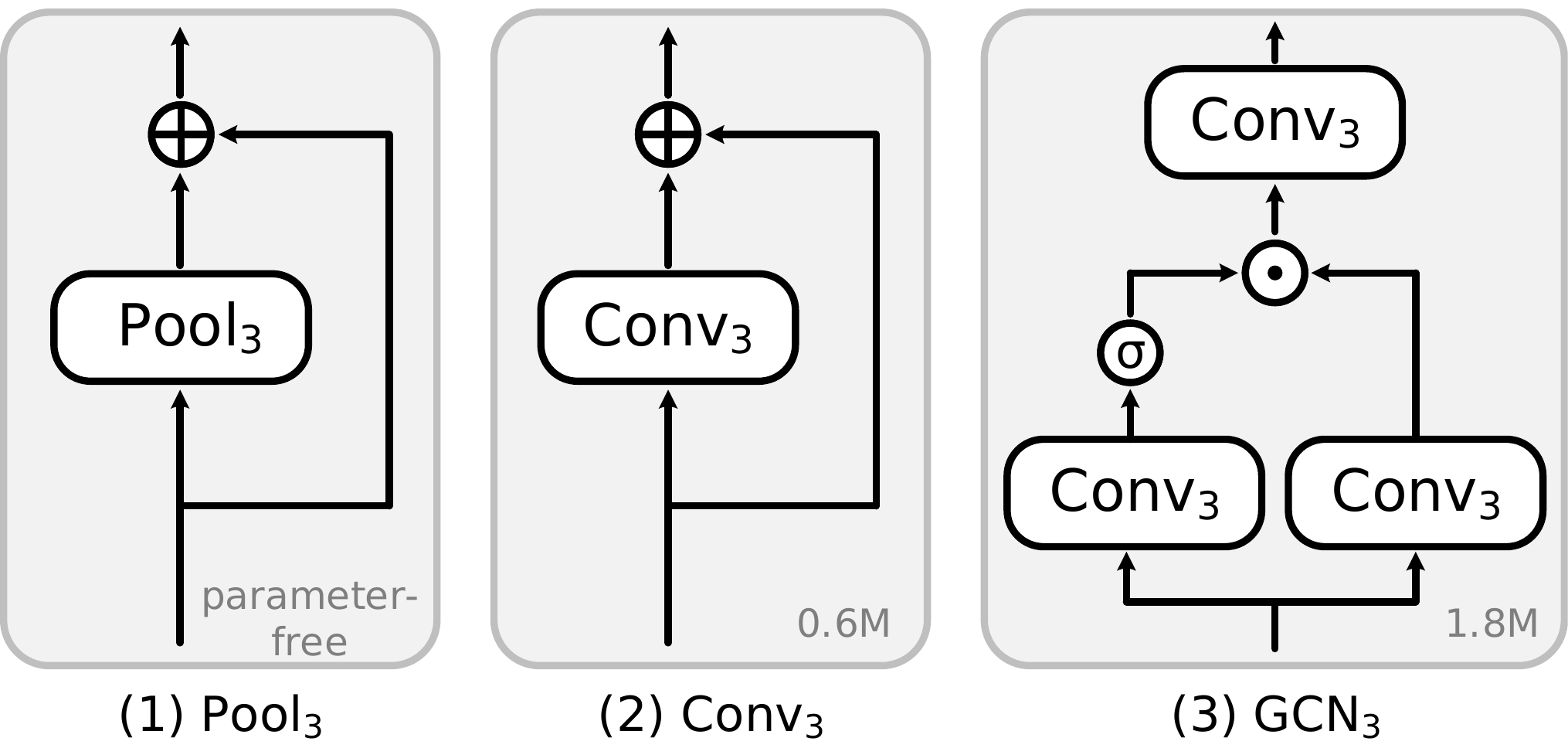}
	\centering
	\vspace{-2mm}
	 \caption{
    \textbf{Three Types of Aggregators.}
	}
	 \label{aggregator}
	 \vspace{-2mm}
\end{figure}

\section{More Experiments}
\label{sec:experiments_supp}
\subsection{Ablation Study on Aggregator}
As shown in \cref{tab:agg_ab_supp} and \cref{erf_agg}, we conduct additional ablation studies on the aggregator.
We evaluate an average pooling layer with a kernel size of $3$, referred to as $\mathrm{Pool}_3$, which represents the simplest parameter-free aggregator.
As illustrated in \cref{erf_agg}, compared to JamMa without an aggregator, $\mathrm{Pool}_3$ extends the effective receptive field to a receptive field spread over the image pair.
As shown in \cref{tab:agg_ab_supp}(1), 
the minimalist aggregator $\mathrm{Pool}_3$ achieves a performance improvement of ($+1.0\%, +0.8\%, +0.9\%$), validating the importance of global dependencies and omnidirectionality.
Further performance gains are observed when learnable parameters are incorporated into the aggregators, as shown in \cref{tab:agg_ab_supp}(2)(3).
Specifically, the gated convolutional unit ($\mathrm{GCN}_3$) improves the performance by ($+2.2\%, +2.2\%, +2\%$).
Additionally, we visualize the effective receptive fields of the models \emph{before training} in \cref{erf_agg}.
Training allows Mamba to establish long-distance dependencies within sequences, while the aggregator extends the sequence dependencies to global dependencies.

\begin{table}[t]
    \centering
    \resizebox{0.8\columnwidth}{!}{
    \begin{tabular}{lccccc} 
    \toprule
    \multirow{2}{*}{Method}    &\multicolumn{3}{c}{Pose est. AUC}\\ 
    \cmidrule(lr){2-4}
     &{@$5^\circ$}  &{@$10^\circ$}  &{@$20^\circ$} \\ 
    \midrule
    w/o Aggregator &62.3 & 75.1 & 84.3 \\
    (1) Pool$_3$ &63.3 &75.9 &85.2\\
    (2) Conv$_3$ &64.2 &76.9 &86.0\\ 
    (3) GCU$_3$ (JamMa) &64.5 &77.3 &86.3\\ 
    \bottomrule
    \end{tabular}
    }
    \vspace{-2mm}
    \caption{\textbf{Ablation Study on Aggregator.}
    }
    \vspace{-3mm}
    \label{tab:agg_ab_supp}
\end{table}

\begin{figure}[t]
	\includegraphics[width=0.99\linewidth]{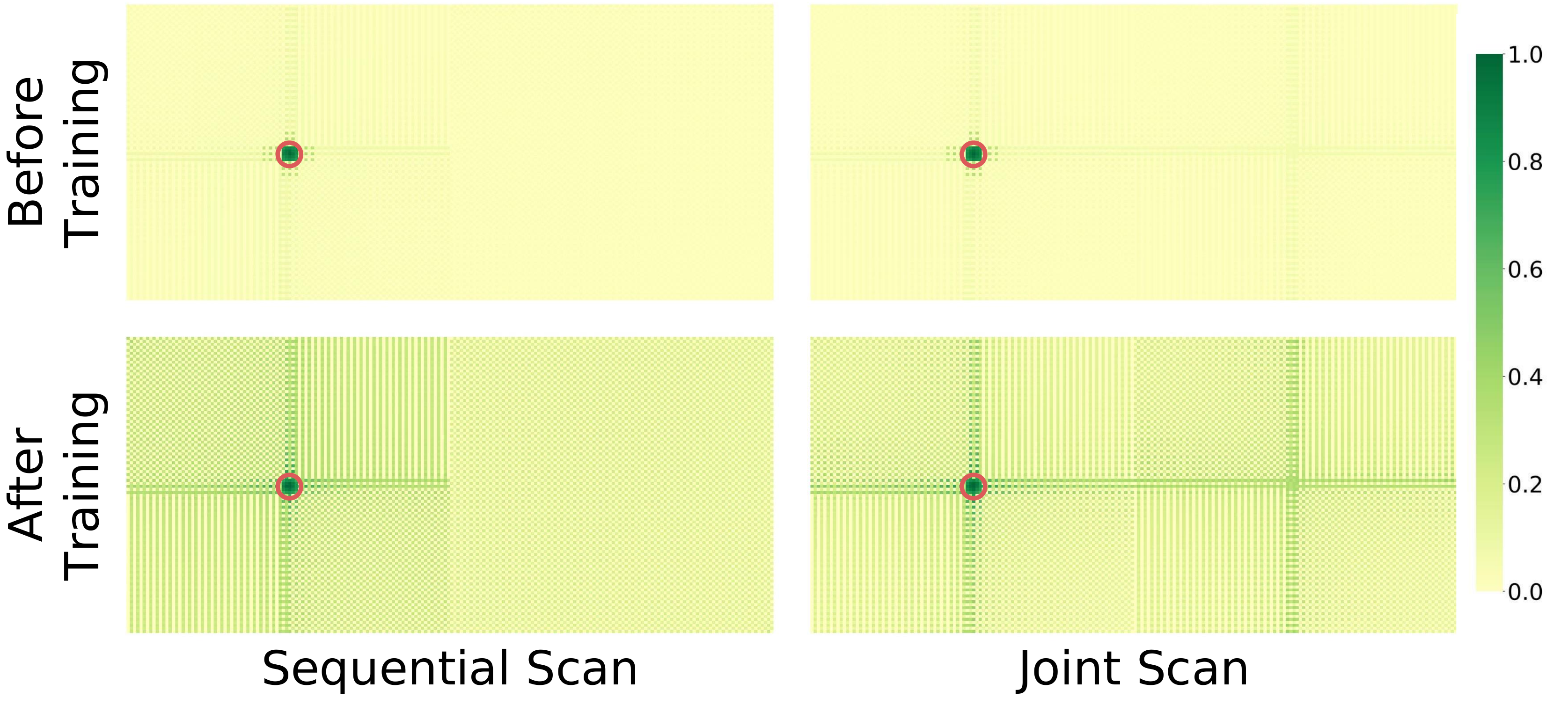}
	\centering
	\vspace{-2mm}
	 \caption{
    \textbf{Effective Receptive Field of Sequential Scan.}
	}
	 \label{erf_ss}
	 \vspace{-4mm}
\end{figure}

\subsection{Effective Receptive Field of Sequential Scan.}
As shown in \cref{erf_ss}, we compare the effective receptive fields of JamMa using sequential and joint scan.
Sequential scan primarily emphasizes internal interactions within a single image but exhibits limited perception of the other image.
In contrast, joint scan enables more comprehensive mutual interactions, 
making it better suited for image matching tasks that require establishing correspondences between \emph{two images}.

\begin{table}[t]
	\centering
	\resizebox{1\columnwidth}{!}{
		\begin{tabular}{ l c c}
			\hline
            \multicolumn{1}{c}{\multirow{2}{*}{Method}} & Day & Night \\
			\cline{2-3}
                     & \multicolumn{2}{c}{$\left(0.25m, 2^\circ\right)$ / $\left(0.5m, 5^\circ\right)$ / $\left(1m, 10^\circ\right)$} \\
			\hline
			DeDoDe$_B$~\cite{dedode} & 87.4 / 94.7 / 98.5 & 70.7 / 88.0 / 97.9    \\
			SP~\cite{superpoint}+LG~\cite{lightglue} & \textbf{89.6} / \textbf{95.8} / \textbf{99.2} & 72.8 / 88.0 / \textbf{99.0}    \\
            LoFTR~\cite{loftr} & 88.7 / 95.6 / 99.0 & \textbf{78.5} / 90.6 / \textbf{99.0} \\
            ASpanFormer~\cite{aspanformer} & 89.4 / 95.6 / 99.0 & 77.5 / \textbf{91.6} / \textbf{99.0} \\
			JamMa & 87.7 / 95.1 / 98.4 & 73.3 / \textbf{91.6} / \textbf{99.0} \\
			\hline
		\end{tabular}
	}
	\vspace{-2mm}
	\caption{\textbf{Visual Localization on the Aachen Day-Night Benchmark v1.1~\cite{aachen}.}}
	\vspace{-2mm}
	\label{Aachen}
\end{table}

\subsection{Visual Localization}
\textbf{Dataset.}
We evaluate our method on the Aachen Day-Night v1.1 benchmark \cite{aachen},
which includes 824 day-time and 191 night-time images selected as query images for outdoor visual localization.

\noindent\textbf{Metric.}
We employ the open-source HLoc pipeline \cite{hloc} for localization and report the percentage of successfully localized images under three error thresholds: $(0.25m, 2^\circ)$, $(0.5m, 5^\circ)$, and $(1m, 10^\circ)$.

\noindent\textbf{Results.}
As shown in \cref{Aachen}, JamMa demonstrates performance comparable to LoFTR and ASpanFormer in outdoor visual localization tasks. 
Note that JamMa is significantly more lightweight, achieving over a $2\times$ reduction in parameters and runtime speedup.

\subsection{Advantage in Low-Resolution Images.}
Quantitative comparisons on low-resolution images are presented in \cref{low_res}.
The results demonstrate that JamMa outperforms ELoFTR by a substantial margin of $+15.8\%$ at a resolution of $256$, while also achieving higher speed.
This highlights JamMa's potential for resource-constrained applications that demand extreme efficiency ($>$$100$ FPS).
The superior performance of JamMa in low-resolution scenarios is attributed to shorter input sequences,
which alleviate the perceptual attenuation issue of Mamba, \ie, the tendency to overlook distant features within a sequence.

\begin{figure*}[t]
	\includegraphics[width=0.99\linewidth]{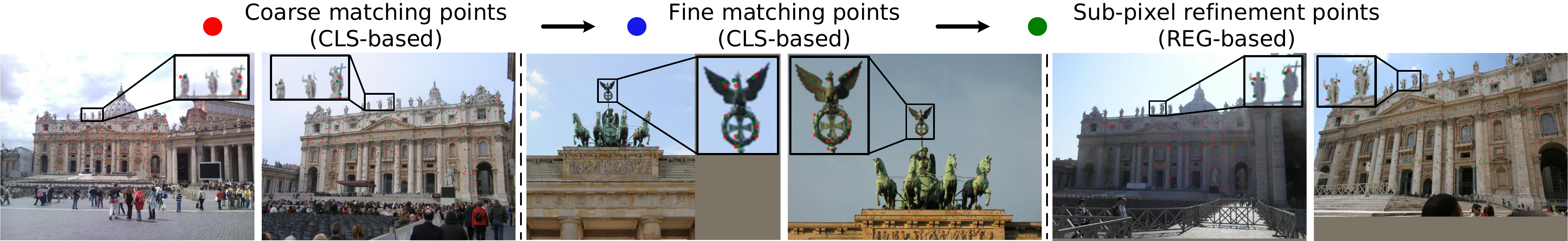}
	\centering
	\vspace{-2mm}
	 \caption{
    \textbf{Visualization of the Coarse-to-Fine Matching Module.} 
	Zoom in for a clearer view.
	}
	 \label{viz_c2f}
	 \vspace{-4mm}
\end{figure*}

\begin{figure}[t]
	\includegraphics[width=0.99\linewidth]{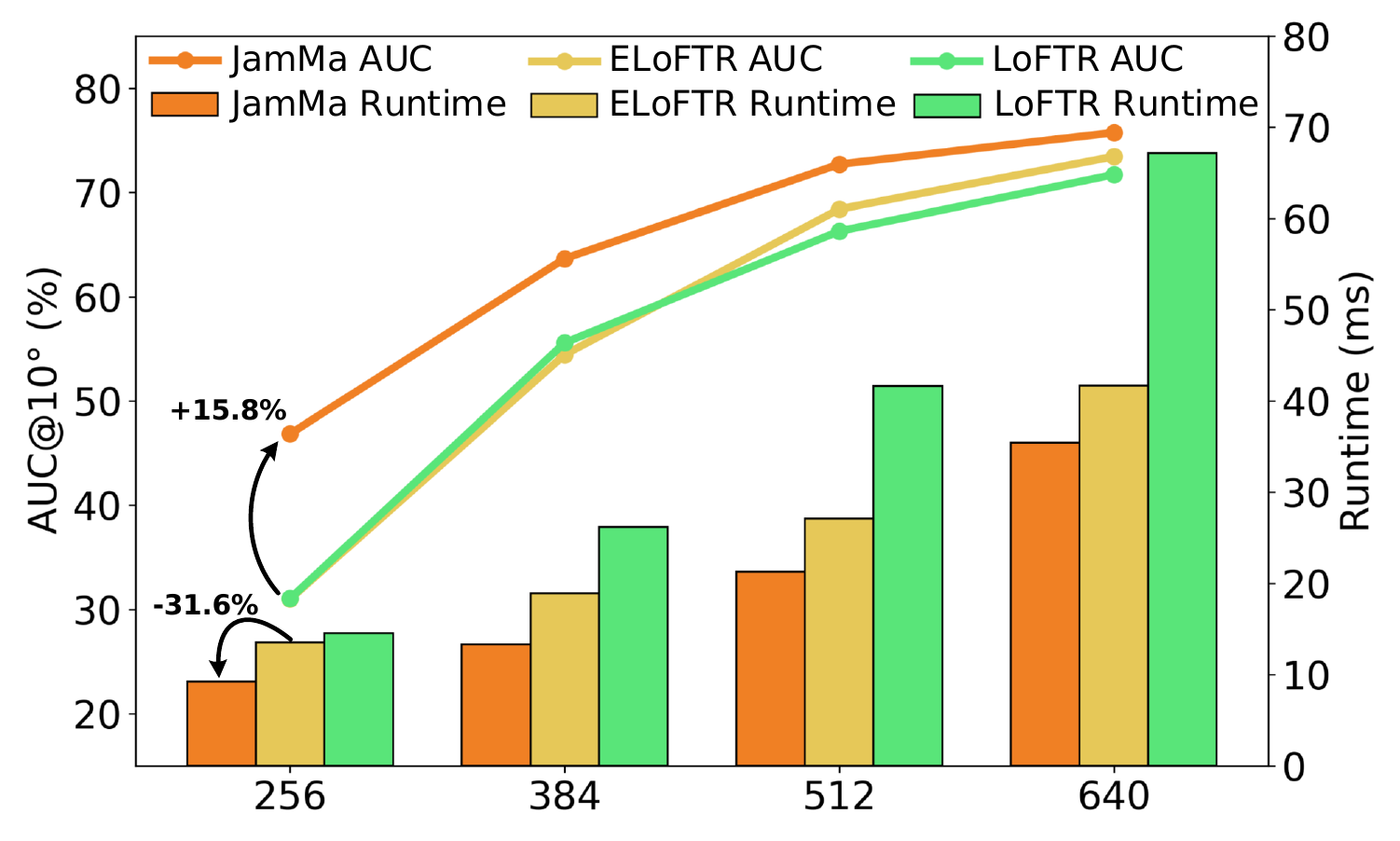}
	\centering
	\vspace{-4mm}
	 \caption{
    \textbf{Comparison in Low-Resolution Images.}
	}
	 \label{low_res}
	 \vspace{-2mm}
\end{figure}

\subsection{Visualization of Coarse-to-Fine Module.}
We adopt the coarse-to-fine matching module proposed in XoFTR \cite{xoftr}, 
which first performs classification-based (CLS-based) coarse matching on coarse grids, 
followed by classification-based fine matching on fine grids, and finally regression-based (REG-based) sub-pixel refinement.
As shown in \cref{viz_c2f}, coarse matching on $1/8$ resolution grids often lacks precision, particularly for small-scale images.
Fine matching on $1/2$ resolution grids significantly enhances precision, while regression-based refinement further improves accuracy to the sub-pixel level.

\begin{table}[t]
    \centering
    \resizebox{1.0\columnwidth}{!}{
    \begin{tabular}{lcccc} 
    \toprule
    \multirow{2}{*}{Method}     & Time   &\multicolumn{3}{c}{Pose est. AUC} \\ 
    \cmidrule(lr){3-5}
    &(ms) &{~@$5^\circ$~}  &{@$10^\circ$}  &{@$20^\circ$} \\ 
    \midrule
    JamMa &61.8 &\textbf{64.5} &\textbf{77.3} &\textbf{86.3}  \\
    (1) w/o sub-pixel ref. &60.2 & 62.1&75.5&84.7 \\
    (2) w/ C2F of LoFTR &\textbf{54.4} & 61.8&74.9&84.4 \\
    \bottomrule
    \end{tabular}
    }
    \vspace{-2mm}
    \caption{\textbf{Ablation Study on Coarse-to-Fine Module.}
    The ref. and C2F denote refinement and coarse-to-fine module, respectively.
    }
    \label{tab:exp ablation_c2f}
    \vspace{-5mm}
\end{table}

\subsection{Ablation Study on Coarse-to-Fine Module}
We evaluate the coarse-to-fine module in LoFTR and the coarse-to-fine module in XoFTR without sub-pixel refinement.
As shown in \cref{tab:exp ablation}(1), sub-pixel refinement supervised by epipolar distance enhances performance by allowing regression-based matching points to achieve sub-pixel accuracy.
Although the coarse-to-fine module in LoFTR is faster, its performance is hindered by two key limitations:
1) one-to-one coarse matching struggles in scenes with significant scale variations, 
and 2) its fine matching does not adjust matching points in the source image.

\subsection{More Qualitative Comparisons}
We provide additional qualitative comparisons of JamMa with LightGlue and ELoFTR in \cref{viz_supp}.
JamMa consistently delivers robust matching results with shorter runtime, achieving lower pose estimation errors.
Further qualitative comparisons for indoor and outdoor scenes are shown in \cref{out_wheel_supp} and \cref{in_wheel_supp}, with matched points color-coded for clarity.

\subsection{Challenging Scenes.}
\begin{figure*}[t]
	\includegraphics[width=0.99\linewidth]{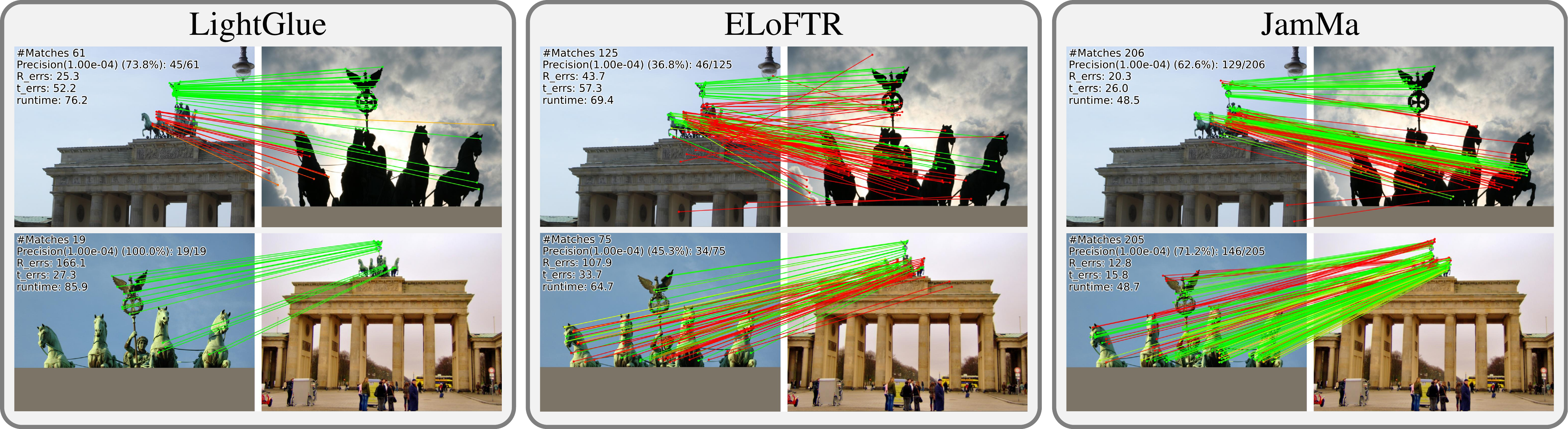}
	\centering
	\vspace{-2mm}
	 \caption{
    \textbf{Challenging Scenes.}
	}
	 \label{viz_chall}
	 \vspace{-2mm}
\end{figure*}

Qualitative comparisons in challenging scenarios are presented in \cref{viz_chall}.
All methods exhibit a significant reduction in the number of matches under drastic illumination and scale variations.
Nevertheless, JamMa maintains a higher number of correct matches, resulting in more robust pose estimation.

\subsection{Failure Cases.}
\begin{figure*}[t]
	\includegraphics[width=0.99\linewidth]{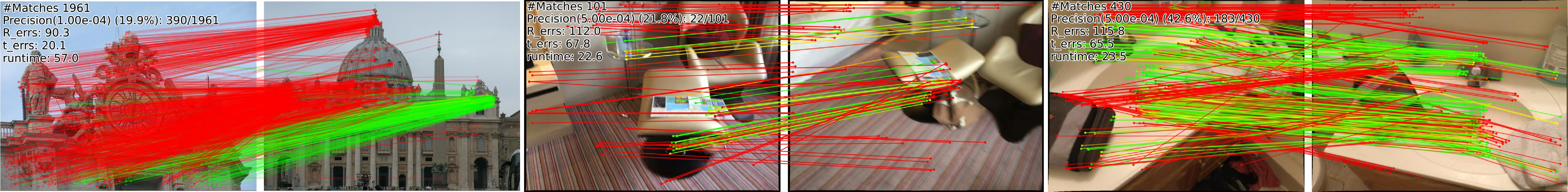}
	\centering
	\vspace{-2mm}
	 \caption{
    \textbf{Failure Cases.}
	}
	 \label{viz_fail}
	 \vspace{-2mm}
\end{figure*}

\begin{figure*}[t]
	\includegraphics[width=0.95\linewidth]{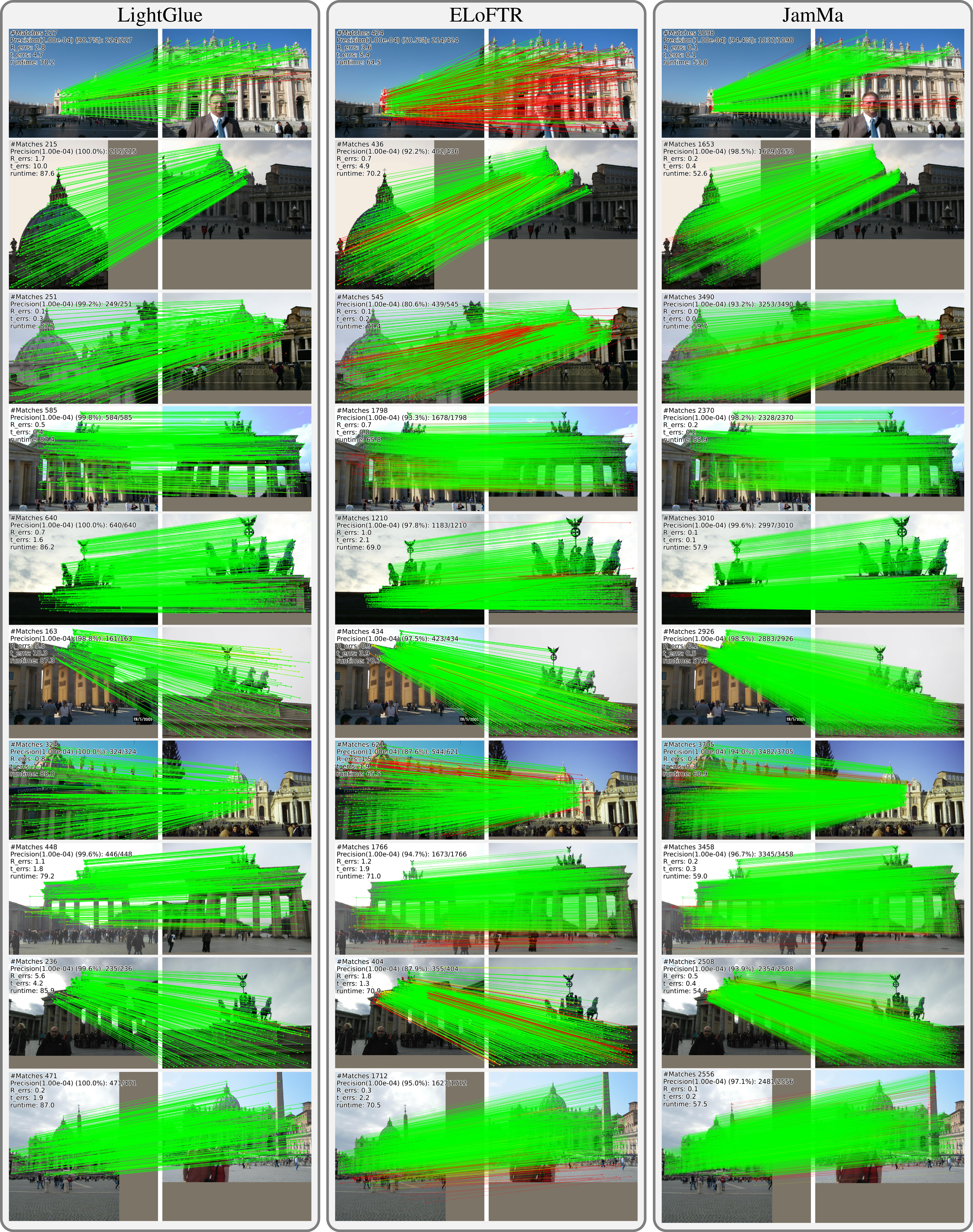}
	\centering
	\vspace{-2mm}
	 \caption{
    \textbf{Comparison of qualitative results.} 
	The reported metrics include precision with an epipolar error threshold of $1 \times 10^{-4}$, rotation and translation errors in pose estimation, and runtime.
	}
	 \label{viz_supp}
\end{figure*}

Failure cases of JamMa are illustrated in \cref{viz_fail}. These cases typically arise in scenarios with extreme scale and viewpoint variations or in texture-less regions.

\begin{figure*}[t]
	\includegraphics[width=0.95\linewidth]{graph/out_wheel_supp.pdf}
	\centering
	\vspace{-2mm}
	 \caption{
    \textbf{Additional qualitative comparisons in outdoor scenes.}
	The matched points are visualized as the same color. 
	}
	 \label{out_wheel_supp}
\end{figure*}

\begin{figure*}[t]
	\includegraphics[width=0.95\linewidth]{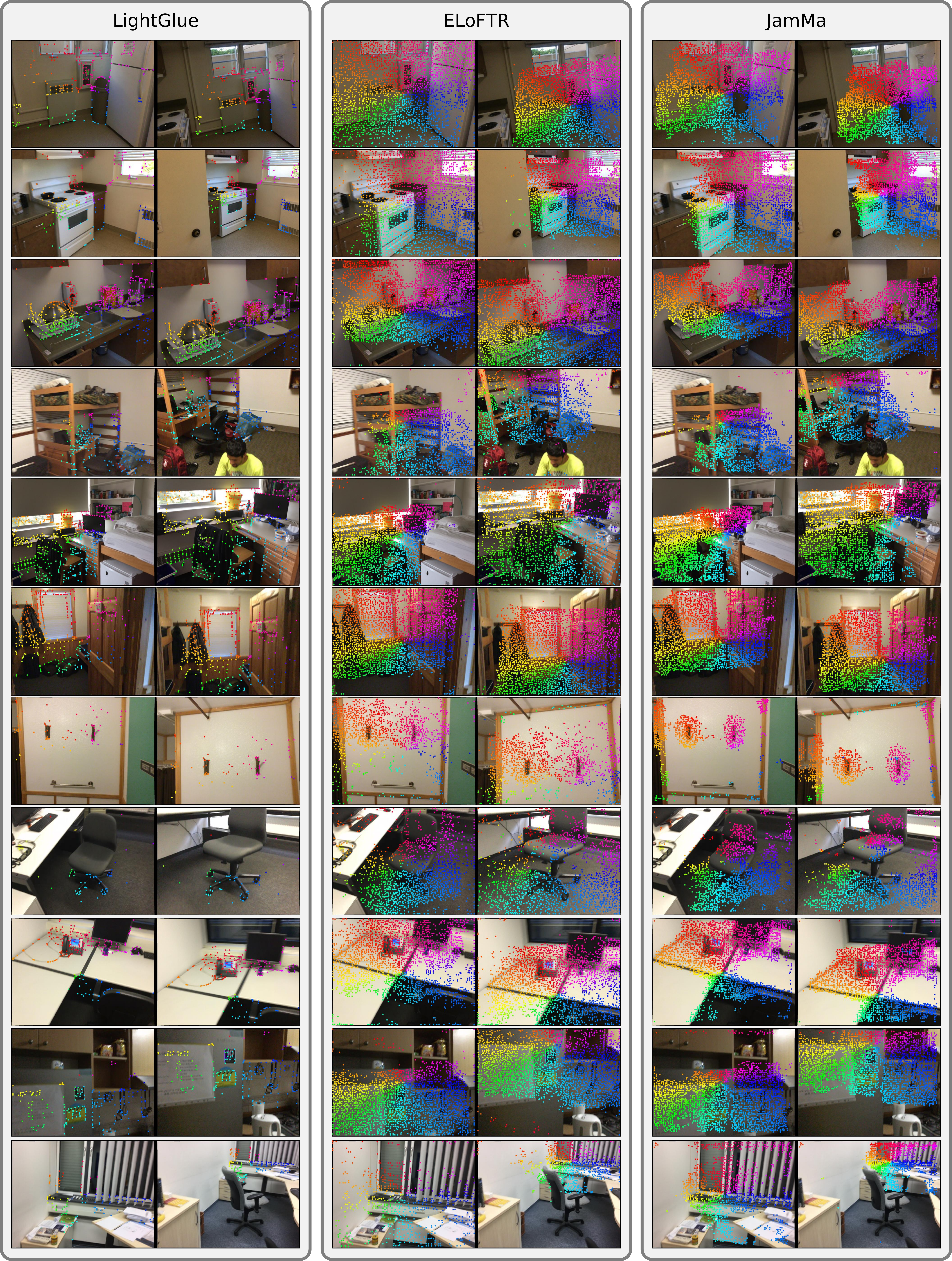}
	\centering
	\vspace{-2mm}
	 \caption{
	\textbf{Additional qualitative comparisons in indoor scenes.}
	The matched points are visualized as the same color. 
	}
	 \label{in_wheel_supp}
\end{figure*}

\end{document}